\documentclass[conference]{IEEEtran}

\IEEEoverridecommandlockouts

\usepackage{tikz}
\usepackage{amsmath}
\usepackage{amsfonts}

\usepackage{filecontents}
\usepackage{graphicx}
\usepackage{float}
\usepackage{flushend}

\ifCLASSINFOpdf
\else
\fi
\begin{document}
%
\title{FedCom: A Byzantine-Robust Local Model Aggregation Rule Using Data Commitment for Federated Learning}

\author{


\IEEEauthorblockN{Bo ~Zhao\IEEEauthorrefmark{1}, Peng ~Sun\IEEEauthorrefmark{2}\IEEEauthorrefmark{3}, Liming ~Fang\IEEEauthorrefmark{1}\IEEEauthorrefmark{4}, Tao ~Wang\IEEEauthorrefmark{1}, and Keyu ~Jiang\IEEEauthorrefmark{1}}
\IEEEauthorblockA{\IEEEauthorrefmark{1}College of Computer Science and Technology, \\Nanjing University of Aeronautics
and Astronautics, NO. 29 Yudao Street, Nanjing 210016, China.\\
\IEEEauthorrefmark{2}School of Science and Engineering, The Chinese University of Hong Kong, Shenzhen, Shenzhen 518172, China\\
\IEEEauthorrefmark{3}Shenzhen Institute of Artificial Intelligence and Robotics for Society (AIRS), Shenzhen 518172, China\\
bozhao@nuaa.edu.cn, sunpeng@cuhk.edu.cn, fangliming1986@163.com, \{wangtao1, jiangky\}@nuaa.edu.cn}

\thanks{
\IEEEauthorrefmark{4}Liming Fang is the corresponding author.
}

%
%
%
}


%


\maketitle

\begin{abstract}

Federated learning (FL) is a promising privacy-preserving distributed machine learning methodology that allows multiple clients (i.e., workers) to collaboratively train statistical models without disclosing private training data. Due to the characteristics of data remaining localized and the uninspected on-device training process, there may exist Byzantine workers launching data poisoning and model poisoning attacks, which would seriously deteriorate model performance or prevent the model from convergence. Most of the existing Byzantine-robust FL schemes are either ineffective against several advanced poisoning attacks or need to centralize a public validation dataset, which is intractable in FL. Moreover, to the best of our knowledge, none of the existing Byzantine-robust distributed learning methods could well exert its power in Non-Independent and Identically distributed (Non-IID) data among clients. To address these issues, we propose FedCom, a novel Byzantine-robust federated learning framework by incorporating the idea of commitment from cryptography, which could achieve both data poisoning and model poisoning tolerant FL under practical Non-IID data partitions. Specifically, in FedCom, each client is first required to make a commitment to its local training data distribution. Then, we identify poisoned datasets by comparing the Wasserstein distance among commitments submitted by different clients. Furthermore, we distinguish abnormal local model updates from benign ones by testing each local model's behavior on its corresponding data commitment. We conduct an extensive performance evaluation of FedCom. The results demonstrate its effectiveness and superior performance compared to the state-of-the-art Byzantine-robust schemes in defending against typical data poisoning and model poisoning attacks under practical Non-IID data distributions.
\end{abstract}


%
\IEEEpeerreviewmaketitle

\section{Introduction}

The conventional machine learning paradigm requires centralizing a massive amount of data from a number of mobile users (i.e., clients) to a central server, which raises serious privacy concerns. The recently emerging federated learning (FL) \cite{Communication-Efficient, Advances_and_Open_Problems} is a promising privacy-preserving distributed machine learning methodology that allows multiple clients to collaboratively train statistical models under the orchestration of a central server while keeping data localized. Typically, the workflow of FL is as follows. The clients download the latest global model from the server. Then, each client performs local model update using her private training data and uploads the updated model parameters to the server. Finally, the server aggregates the received information and updates the global model. The above steps are repeated until the global model converges. FL could not only construct high-quality machine learning models but could also significantly mitigate privacy risks, thus attracting significant attention recently \cite{FederatedLearninginDistributedMedicalDatabases, PrivacyPreservingTextRecognition}.

However, since the server neither has access to clients’ private training data nor could inspect the distributed on-device training process in FL, there usually exist Byzantine workers launching data poisoning \cite{ManipulatingMachineLearning:Poisoning, TowardsPoisoningofDeepLearningAlgorithmswithBack-gradient, UnderstandingDistributedPoisoning, Min-MaxOptimizationwithoutGradients:ConvergenceandApplications} and model poisoning \cite{LocalModelPoisoningAttacksto, HowToBackdoorFederatedLearning, DistributedBackdoorAttacksagainstFederated} attacks. Data poisoning attacks refer to fabricating or falsifying training samples, which induces an abnormal training process and leads to poisoned local models. Model poisoning attacks represent that Byzantine workers directly modify the local model updates. It is worth noting that the essence of these two kinds of attacks is to generate a poisoned local model and make it be aggregated when the server updates the global model. Without effective strategies for defending against these attacks or Byzantine-robust FL schemes, the global model’s performance would drastically deteriorate, or the global model would not converge. To address this issue, researchers have proposed several Byzantine-robust FL approaches. Existing Byzantine-robust FL schemes could be generally grouped into two categories. We name one \cite{Machine_Learning_with_Adversaries, TheHiddenVulnerabilityofDistributed, Byzantine-RobustDistributedLearning:Towards, Byzantine-RobustStochasticAggregationMethodsfor} as Distance Statistical Aggregation (DSA), as such schemes assume that poisoning local models obtain great distances from other benign local models, and remove statistical outliers before using them to aggregate a global model. We name the other \cite{Zeno:DistributedStochasticGradientDescent, Zeno++:RobustFullyAsynchronous} as Contribution Statistical Aggregation (CSA), as such schemes give priority to local models which contribute the most to global model's accuracy or global model's convergence. The basic principle of these two kinds of methods is that instead of performing model averaging when updating the global model, the server would first try to identify the poisoned local model updates and exclude them from local models aggregation or perform quality-aware local model aggregation.

Nevertheless, existing Byzantine-robust FL schemes have performance limitations from the following three perspectives. First, the DSA mechanisms work under the strict assumption that the poisoned local models submitted by Byzantine workers are statistically far away from the local models from benign workers. However, the recently proposed attack \cite{LocalModelPoisoningAttacksto} could generate poisoned local models statistically close to benign models and could circumvent the DSA schemes. Second, the CSA methods require collecting a public validation dataset before the training starts, which is intractable and unpractical in FL, as clients are unwilling to share their private data due to privacy concerns. Finally, most of the existing Byzantine-robust FL schemes are ineffective when data is non-independent and identically distributed (Non-IID) among clients, which is usually the case in reality as different users bear different device usage patterns. The reason is that the local models updated based on Non-IID data among different clients are pretty different from each other, making it even more challenging to distinguish poisoned local models from benign ones.

To address the issues mentioned above, we propose FedCom in this paper. FedCom is a novel Byzantine-robust federated learning framework, which could achieve both data poisoning and model poisoning tolerant FL under practical Non-IID data distribution among clients, by incorporating the idea of commitment from cryptography. The core idea of commitment is "blind verification": integrity of localized data or process could be verified without disclosing any of it. If there is an effective mean of committing local data's distribution in FL, then we could indirectly inspect local datasets or local training processes. Specifically, in FedCom, each client is first required to make a commitment to its local training data distribution. In more detail, data commitment is an artificial dataset which takes the average of a sample cluster in original dataset as a sample. The data commitment constructed in this way is empirically demonstrated to subject to the similar data distribution to the original dataset, without disclosing any samples in it. Then, the central server identifies poisoned datasets by comparing the Wasserstein distance among commitments submitted by different clients. This is because an effective data poisoning attacks usually manipulate the data distribution, and there may obtain significant Wasserstein distance between manipulated distribution and benign one. Furthermore, we consider obvious asynchronous progress of a local model's convergence relative to others as dishonesties, and distinguish them from benign ones via testing each local model on its corresponding data commitment, and accordingly adjust the weights of different local models when conducting local models aggregation at the server.

In summary, the major contributions of this paper are summarized as follows:


\begin{itemize}
    \item To the best of our knowledge, this is the first Byzantine-robust FL framework that could simultaneously defend against both data poisoning and model poisoning attacks in practical application scenarios with non-IID data distribution among clients. In particular, we adapt the idea of commitment from cryptography to FL and propose FedCom, a novel Byzantine-robust FL scheme based on data commitment.
    \item We identify poisoned local datasets and thus thwart poisoned local models induced by data poisoning attacks by measuring the Wasserstein distance from each data commitment to the remaining ones and scoring each data commitment according to the measured distance valuation.
    \item We identify dishonest local training processes and eliminate crafted poisoning local models by testing each local model on its corresponding data commitment and banning models with significantly asynchronous convergence processes from aggregation.
    \item We conduct an exhaustive performance evaluation of FedCom. Specifically, we make extensive comparisons with the state-of-the-art Byzantine-robust FL schemes defending against two typical data poisoning attacks and two typical model poisoning attacks based on two non-IID datasets. The results demonstrate the effectiveness and the superior performance of the proposed FedCom.
\end{itemize}
\textbf{Organization of paper}: in Section \ref{s2}, we introduce preliminaries and attack model of our work, in Section \ref{s3}, we introduce the design of data commitment and FedCom. Section \ref{s4} is the evaluation of our work, and Section \ref{s5} and \ref{s6} are related work and conclusion of our work.

\section{Preliminaries and Attack Model}
\label{s2}
\subsection{Federated Learning}
Federated learning (FL) is a novel privacy-preserving distributed learning framework that allow multiple \emph{workers} to collaboratively train a machine learning model without disclosing any private data. FL require a \emph{central server} to coordinate workers' training processes. In each iteration of FL, central server should firstly assign the latest global model to each worker. Secondly, each worker should update this model locally with its local dataset and obtain a \emph{local model}. Then, each worker should submit its local model to central server. Finally, the central server collects local models from involved workers, and aggregate them with a certain \emph{aggregation rule} to obtain the latest global model in next iteration. Iteration repeats until the global model converges to a desired accuracy, or the number of iteration reaches a certain threshold. The vanilla FL system adopt \texttt{FedAverage} as the aggregation rule. Concretely, an iteration under \texttt{FedAverage} follow steps below:

\textbf{Step 1}: the central server distribute latest global model $W$ to each worker.

\textbf{Step 2}: the $i$th worker trains local model $w_{i}$ locally, and submit it to central server.

\textbf{Step 3}: central server obtain the new global model $W=\sum_{i=0,1, \ldots n}\frac{|d_{i}|}{\sum_{j=0,1, \ldots n}|d_{j}|}w_{i}$, where $|d_{i}|$ is the size of local dataset $i$th worker holds.

A critical step in FL is how the server aggregates the local model updates from different clients to conduct global model updates. \texttt{FedAverage} has been demonstrated empirically to perform well in terms of model accuracy when nothing abnormal happens. However, since the server neither has access to workers' private training data nor could inspect the distributed on-device local training process, its common that there exist Byzantine workers launching poisoning attacks, which will significantly degrade the effectiveness of \texttt{FedAverage} \cite{Machine_Learning_with_Adversaries}. Thus, it is necessary to develop and employ some Byzantine-robust aggregation methods to ensure satisfactory model performance.

\subsection{Attack Model}
To introduce our work, we need to model the attackers in FL, and design our scheme specifically.

\textbf{Attacker's method}: In this work, we mainly consider two kinds of Byzantine attacks, i.e., \emph{local dataset poisoning attacks} (LDP) and \emph{local model poisoning attacks} (LMP). Such two categories of attacks are aimed to prevent global model's convergence, or manipulate global model. LDP or LMP requires attacker to submit poisoned local models. These models will be aggregated into global model if attack succeed, and manipulated global model will fit a poisoned data distribution $\mathbb{E}(D_{poisoning})$, where $D_{poisoning}$ is poisoned dataset. $\mathbb{E}(D_{poisoning})$ should be different from the original distribution $\mathbb{E}(D_{training})$, and model trained on which may not convergence, or could be triggered incorrect inference. For example, Error-Generic Poisoning Attacks and Error-Specific Poisoning Attacks \cite{TowardsPoisoningofDeepLearningAlgorithmswithBack-gradient} train models which fits a poisoned distribution and obtain maximum loss on validation set, while backdoor attack \cite{HowToBackdoorFederatedLearning} trains models which fits both benign distribution and a specific backdoor distribution (it could possibly contains slight difference from the original distribution), and backdoor distribution could trigger wrong inference. As to LMP, it attempts to directly craft poisoning local models to make $W$ fit an unknown distribution.

\textbf{Attacker's resources}: Similar to many existing works, we assume that the number of Byzantine workers is less than half of the total number of workers in the system. Moreover, we assume that Byzantine works could collude to attack, which makes it more challenging for the server to detect these attacks.


\textbf{Attacker's knowledge}: In this paper, we consider an ideal case for Byzantine worker. In other words, Byzantine workers own full knowledge of the aggregation rule $A$ and benign local models from benign workers. It is worth noting that this could make the benign global model more easily estimated by Byzantine workers, hence causes the most significant difficulty for robust FL.


\section{Design of FedCom}
\label{s3}
In this section, we elaborate on the design details of the proposed robust FL scheme FedCom. Specifically, we first present the idea of data commitment by introducing a toy example. Then, we illustrate how to defend against local model poisoning attacks based on the submitted commitment. Finally, we present the detailed design on how to thwart local data poisoning attacks. We would like to point out that throughout the design of FedCom, we focus on a practical case, where the data among workers is Non-IID.

\subsection{Detailed Design of Data Commitment}

Under condition that local datasets are Non-IID, common assumption that benign local models are statistically similar is disabled. Hence, the most effective mean of defending poisoning attacks in FL could be inspecting local datasets and local training processes, without directly insulting user's privacy. Recall common cryptographic tools like hash function \cite{Multi-CollisionResistantHashFunctions} and consider the game below: Alice and Bob are playing guessing game, in which player gives a bigger number wins the game. Assume that the number is public once given, and we can't ensure two players giving their numbers simultaneously. Obviously, once a player gives the first number, the other player could simply win the game by giving the second number bigger than the first one. We could use hash function to fix the bug: both two players should firstly give a hash digest of the numbers they are giving, and give the original number after two digests are given. The hash digest will not match if anyone changed the original number after giving the digest, and it is almost impossible for other players to infer the original number from hash digest. As a result, we have a simple way to check if players have changed the numbers without knowing them. This is the basic idea of commitment system \cite{OntheCompositionofTwo-ProverCommitments} in cryptography. Back to our task, if we could construct commitments for local datasets without disclosing any sample, and verify local training processes using such commitments, then we could provide an indirect mechanism to detect dishonest local training process. In such a mechanism, whether a local model is trained honestly could be only verified by corresponding commitment, which may have no concern with Non-IID between local datasets.

Recall the definition of LMP we proposed above: honest workers submit benign local models $w^{(b)}_{i}$, while a Byzantine worker submits crafted poisoning model $w^{(p)}_{i}$, $W=A(w^{(p)}_{1}, w^{(p)}_{2}, ..., w^{(p)}_{f}, w^{(b)}_{1}, w^{(b)}_{2}, ..., w^{(b)}_{n-f})$ fits an unknown data distribution. The essence of such attacks is that Byzantine workers dishonestly train local models on its benign local dataset $D^{(i)}_{training}$. To detect such dishonesty, we need a certain method to verify the local training process of workers.

We could neither directly access to $D^{(i)}_{training}$ of $worker_{i}$, nor check its local training process. However, we could ask $worker_{i}$ to submit the description of $\mathbb{E}(D^{(i)}_{training})$, which we call \emph{commitment}. Such commitment should 1) reflects $\mathbb{E}(D^{(i)}_{training})$ as accurate as possible, 2) not disclose any original samples in $D^{(i)}_{training}$, 3) be able to verify $w_{i}$ whether it is honestly trained on $D^{(i)}_{training}$. Formally,

\begin{equation}
\begin{aligned}
&D^{(i)}_{commitment} \sim \mathbb{E}(D^{(i)}_{training}) \\
s.t.&\quad D^{(i)}_{commitment} \cap D^{(i)}_{training}=\emptyset.
\end{aligned}
\label{(1)}
\end{equation}

In conclusion, we define ideal commitment $D^{(i)}_{commitment}$ is a dataset that subjects to the same distribution of $D^{(i)}_{training}$, but have no intersection with it. Hence, we consider how to construct $D^{(i)}_{commitment}$. Assume that $x^{(i)}_{j}$, the $j$th dimension of all samples in $D^{(i)}_{training}$, subjects to a certain Gaussian distribution, formally $x^{(i)}_{j} \sim N(\mu^{(i)}_{j}, (\sigma^{(i)}_{j})^{2})$. Obviously, the average of $x^{(i)}_{j}$ could reflect the partial features of $N(\mu^{(i)}_{j}, (\sigma^{(i)}_{j})^{2})$. However, the process of average have lost some key information like $(\sigma^{(i)}_{j})^{2}$, or $D^{(i)}_{commitment}$'s edge geometrical features in high-dimension space. To maintain features above, the process of average should be more "fine-grained". For each sample $p_{k}$ in $D^{(i)}_{training}$, we firstly find the $m$ nearest samples $\Gamma=\{p_{k+1}, p_{k+2}, ..., p_{k+m}\}$ to $p_{k}$. Then, take $c_{j}=\frac{1}{m}\sum_{p \in \Gamma} p$ as crafted sample $c_{k}$. Traversing all samples in $D^{(i)}_{training}$ , and we could finally construct a set of crafted samples $\{c_{1}, c_{2}, ..., c_{|D^{(i)}_{training}|}\}$ as commitment of $D^{(i)}_{training}$.

\subsection{Defending Against LMP Attacks}

We could refer to CSA to verify a worker's local training process using its commitment. Concretely, we also calculate the difference of $w_{i}$'s loss on $D^{(i)}_{commitment}$ between contiguous iterations in FL to decide the weight of $w_{i}$ in aggregation. It should be noted that process mentioned above is essentially different from CSA, as a public validation set measures $w_{i}$'s contribution to global training process, while $D^{(i)}_{commitment}$ measures the progress of $w_{i}$ honestly trained on corresponding $D^{(i)}_{training}$.

Concretely, let $l_{i}=l^{before}_{i}-l^{after}_{i}$, in which $l^{before}_{i}$ is $w_{i}$'s loss on $D^{(i)}_{commitment}$ or public testing set before single round local training, while $l^{after}_{i}$ is loss after single round local training. In CSA, a smaller $l_{i}$ reflects a smaller $w_{i}$'s contribution to global training. However, in FedCom, we consider a smaller $l_{i}$ as a reflection of high level convergence of $w_{i}$. $l_{i}$ must be positive if hyperparameters are well-tuned, or we consider $w_{i}$ is not honestly trained on $D^{(i)}_{training}$ in current round local training. We believe that with the same machine learning task, hyperparameters and model structure, each local model's convergence should be approximately synchronous. In other word, differences between $l_{i}$ should not be too large. Hence, our strategy of defending LMP is to filtering out local models with negative $l$s, and give priority to local models with $l$ which is closest to other $l$s. Concretely, in $k$-worker FL system, the scoring rule of $w_{i}$'s \emph{Training Credit} (TC) is

\begin{equation}
T C_{i}=\left\{\begin{array}{c}
0, l_{i} \leq 0 \\
\frac{1}{\sum_{l_{j}>0}\left|l_{i}-l_{j}\right|}, l_{i}>0
\end{array}\right.
\label{(2)}
\end{equation}

Obviously, $w_{i}$ with a negative $l$ will receive a TC score of zero, and such $w_{i}$ will be abandoned by central server. $l$ with larger distances to other $l$s brings a lower TC score to corresponding local model, as such a local model's local training progress may not synchronized to others.

Such a scheme could detect LMP in FL from the underlying level, as $w^{(p)}_{i}$ is usually crafted along the deviated update direction of benign local models, and $w^{(p)}_{i}$ with significant poisoning effect usually obtains worse performance on benign validation dataset. Hence, if $D^{i}_{commitment} \sim \mathbb{E}(D^{i}_{training})$, then $w^{(p)}_{i}$ should also obtain worse performance on $D^{i}_{commitment}$. Such scheme does not disclose any original training samples, as samples in $D^{i}_{commitment}$ are crafted by averaging sample clusters in $D^{i}_{training}$, and the reverse process could be difficult.

This scheme is still a simple scheme due to obvious limitations, as such scheme could not defend LDP, as $D_{commitment}$ could not directly reflect if $\mathbb{E}(D_{training})$ is honest or not. If $D^{i}_{commitment}$ is crafted based on $D^{i}_{poisoning}$, then poisoning local model could still have positive $l$ on $D^{i}_{commitment}$. Hence, we need to propose a scheme to estimate $\mathbb{E}(D^{i}_{training})$'s honesty.

\subsection{Defending Against LDP Attacks}

In FL, directly estimating the honesty of $D^{i}_{commitment}$ and corresponding $D^{i}_{commitment}$ could be difficult, as we introduced in Section \ref{s2}, it is difficult to collect a global validation dataset as benchmark. Hence, central server could only estimate via limited knowledge, which is local models and corresponding commitments.

\begin{figure}[t]
\centering
\includegraphics[width=0.48\textwidth]{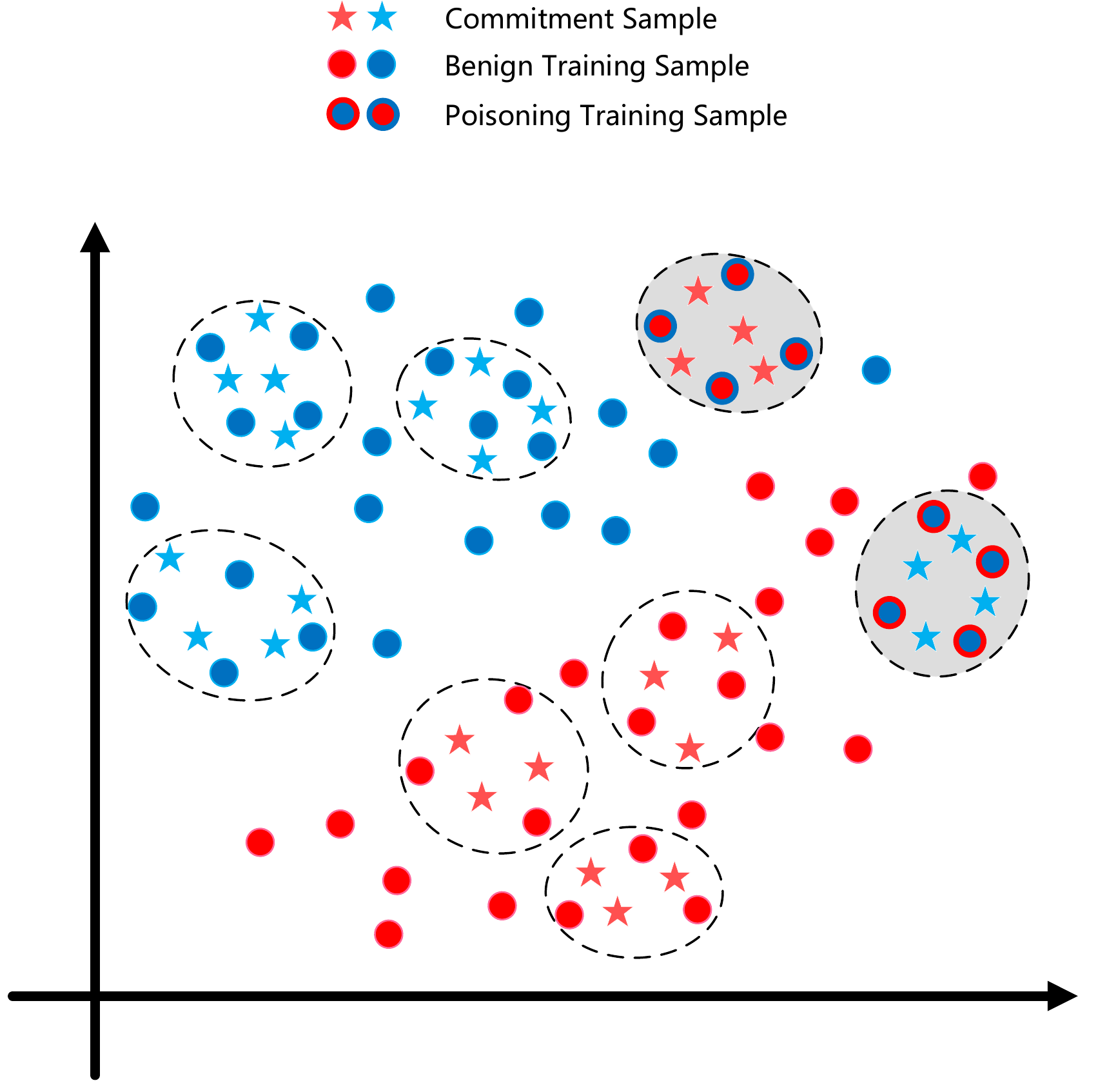}
\caption{Demonstration of data commitment. Data commitment does not disclose any original training samples, but subjects to the same distribution to original training dataset. Obviously, poisoning local datasets (gray circles) obtain data commitment that is geometrically "far away" from other data commitments (white circles), hence obtain a low DC. Obviously, a model's performance on original local datasets should be close to that on corresponding commitment datasets, otherwise model may not fit the original local dataset, hence obtain a low TC.}
\label{commitment}
\end{figure}

Obviously, there are significant differences between commitments in Non-IID data circumstances of FL. However, we could still search for rules among such differences. Although local datasets subject to different data distributions, but there is a rational assumption that if $D^{(i)}_{training}$'s partition is random, then the difference between local datasets, as well as corresponding $D^{(i)}_{commitment}$, obtain a certain upper bound, which is determined by the machine learning task. For example, the dissimilarity between two facial images from different users could be significant, but may not be more significant than dissimilarity between a human facial image and an aeroplane image. Hence,we assume that the dissimilarities between $D^{(i)}_{training}$ subject to a certain Gaussian distribution, as well as dissimilarities between $D^{(i)}_{commitment}$.

In this work, we adopt \emph{Wasserstein distance} \cite{FromGANtoWGAN} to measure the dissimilarities between $D^{(i)}_{commitment}$, as Wasserstein distance could define distance between two datasets from the aspect of each single sample. Moreover, Wasserstein distance could apply to two datasets without overlap compared with common measurement like KL divergence. This is of significant importance to Non-IID data circumstance in FL, as local datasets distributions may not "overlap" with high degree of Non-IID. Firstly, we need to measure the distances between one commitment and others from the perspective of dimensions. We take $x_{j}$ of all $D_{commitment}$ as universal set, then the supplementary set of $x^{(i)}_{j}$ in $D^{(i)}_{commitment}$ is $\overline{x^{(i)}_{j}}=\{p_{k}|p_{k}\in x^{(0)}_{j}, x^{(1)}_{j}, ..., x^{(i-1)}_{j}, x^{(i+1)}_{j}, ..., x^{(k)}_{j}\}$. Then, we adopt the Wasserstein distance between $x^{(i)}_{j}$ and $\overline{x^{(i)}_{j}}$ of $D^{(i)}_{commitment}$ to measure the difference between $D^{(i)}_{commitment}$ and other commitments on \emph{j}th dimension. Finally, we obtain the total distribution divergence between $D^{(i)}_{commitment}$ and other commitments. Concretely, in \emph{n}-dimension FL system, the scoring rule of $w_{i}$'s \emph{Data Credit} (DC) is

\begin{equation}
\begin{array}{l}
D C_{i}=1-\frac{1}{\sqrt{2 \pi} \sigma} \int_{-\infty}^{d^{(i)}} e^{-\frac{(t-\mu)^{2}}{2 \sigma^{2}}} d t \\
d^{(i)}=\frac{1}{n} \sum_{j \in 1,2, \ldots, n} Wass\left(x^{(i)}_{j}, \overline{x^{(i)}_{j}}\right)
\end{array}
\label{(3)}
\end{equation}

in which $d^{(i)}$ is the total distribution divergence of $D^{(i)}_{commitment}$, and $\mu$ and $\sigma$ are the mean and variance of $d$ of each commitment. Scoring rule mentioned above directly reflects the distribution divergence between each local dataset and global dataset in high dimension space. Obviously, assuming that $d^{(i)}$ subjects to a certain Gaussian distribution, a smaller divergence will lead to a higher DC. The core idea of commitment could be demonstrated in Figure \ref{commitment}.


\subsection{Combination of Data Credit and Training Credit}

Obviously, DC and TC are separately used to defend LDP and LMP. Hence, to defend such two types of attacks simultaneously, we need to adopt an effective combination of DC and TC. We need to construct two positive correlations between the final score and DC or TC. Any descent of DC or TC will lead to the descent of a local model's final score. In conclusion, we define the final score of a local model as the product of DC and TC.

Meanwhile, we consider two adjustments. Firstly, in order to eliminate negative effect of poisoning local models, once we detect a poisoning model, we need to ban this model from global model aggregation. Concretely, in $k$-worker FL system, we calculate the final score of each local model, and set the weight of local models with smaller final scores than median as zero. Secondly, to keep \texttt{FedAverage}'s generality to Non-IID local datasets, we need to adjust the weight of benign local models according to the size of local datasets. Concretely, in $k$-worker FL system, the final weight $I_{i}$ of $w_{i}$ is determined that

\begin{equation}
I_{i}=\frac{{Flag}_{i} \&\left|D^{(i)}_{\text {training }}\right|}{\sum_{j=1,2, \ldots, k} {Flag}_{j} \&\left|D^{(j)}_{\text {training }}\right|} \\
\label{(4)}
\end{equation}

in which

\begin{equation}
{ Flag }_{i}=\left\{
\begin{array}{l}
0, D C_{i} * T C_{i}<\operatorname{Median}(D C * T C) \\
1, D C_{i} * T C_{i} \geq \operatorname{Median}\left(D C* T C\right)
\end{array}\right.
\label{(5)}
\end{equation}

Following processes of determining TC and DC mentioned above, each local model under FedCom could obtain a final weight.

\subsection{Brief Summary}
Before starting the process of FL, each worker should submit $D^{(i)}_{commitment}$, and its DC will be determined. After that, each update of $w_{i}$ will be validated on $D^{(i)}_{commitment}$ to determine TC. In general, FedCom perform \texttt{FedAverage} with a strategy of filtering out poisoning models, so that global model will be protected from attack of poisoning local models, and obtain well performance on Non-IID data circumstances as well. Following processes we design, FedCom could achieve both data poisoning and model poisoning tolerant FL under practical Non-IID data partitions.

\section{Evaluation}
\label{s4}

In this section, we conduct an exhaustive performance evaluation of FedCom and make comparisons with several Byzantine-robust FL schemes in terms of the performance in defending against four typical poisoning attacks.

\subsection{Experimental Settings}

\begin{figure*}[t]
\centering
\begin{minipage}[t]{0.24\textwidth}
\centering
\includegraphics[width=4.8cm]{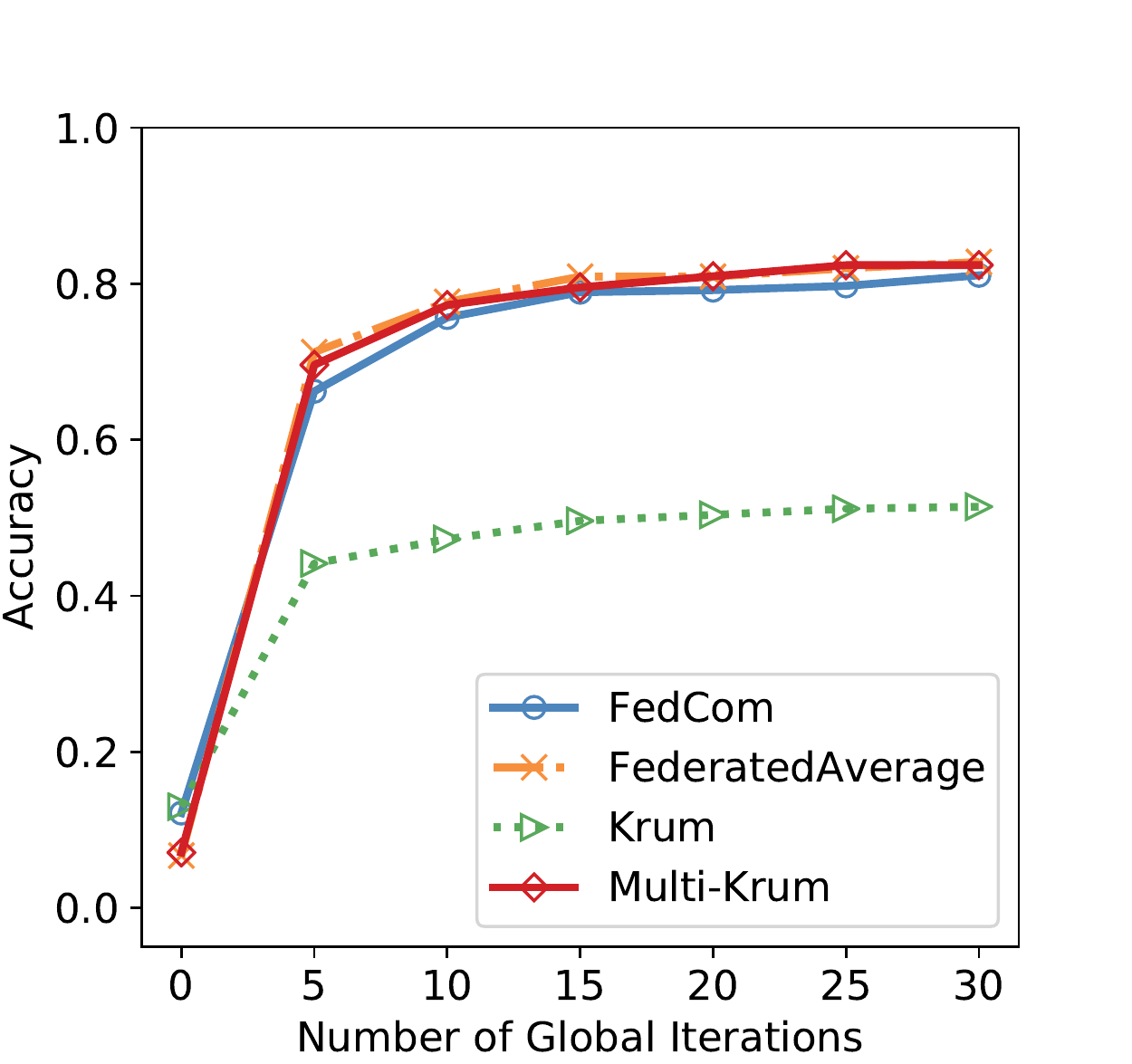}
\centerline{(a) MNIST}
\end{minipage}
\begin{minipage}[t]{0.24\textwidth}
\centering
\includegraphics[width=4.8cm]{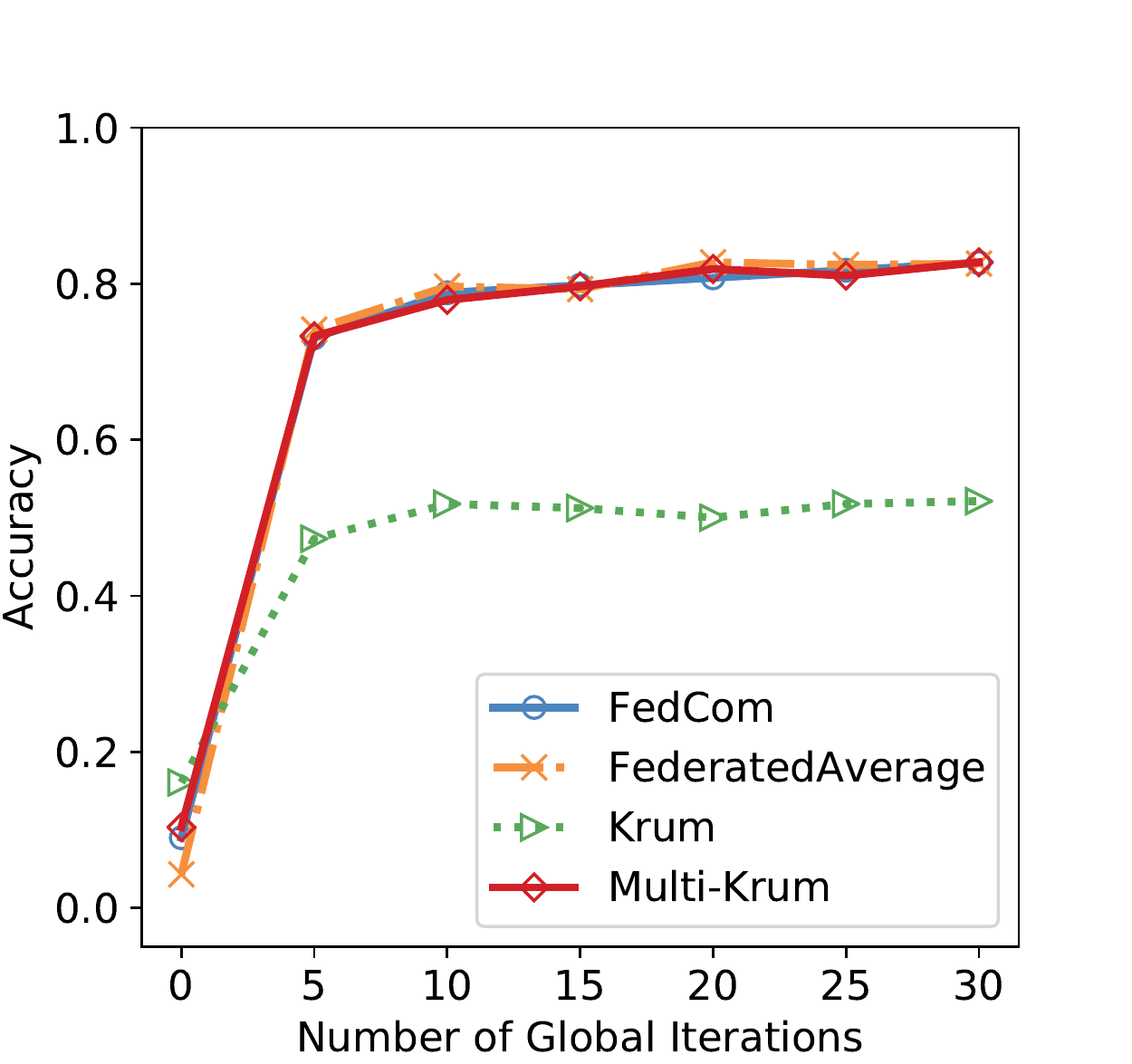}
\centerline{(b) MNIST}
\end{minipage}
\begin{minipage}[t]{0.24\textwidth}
\centering
\includegraphics[width=4.8cm]{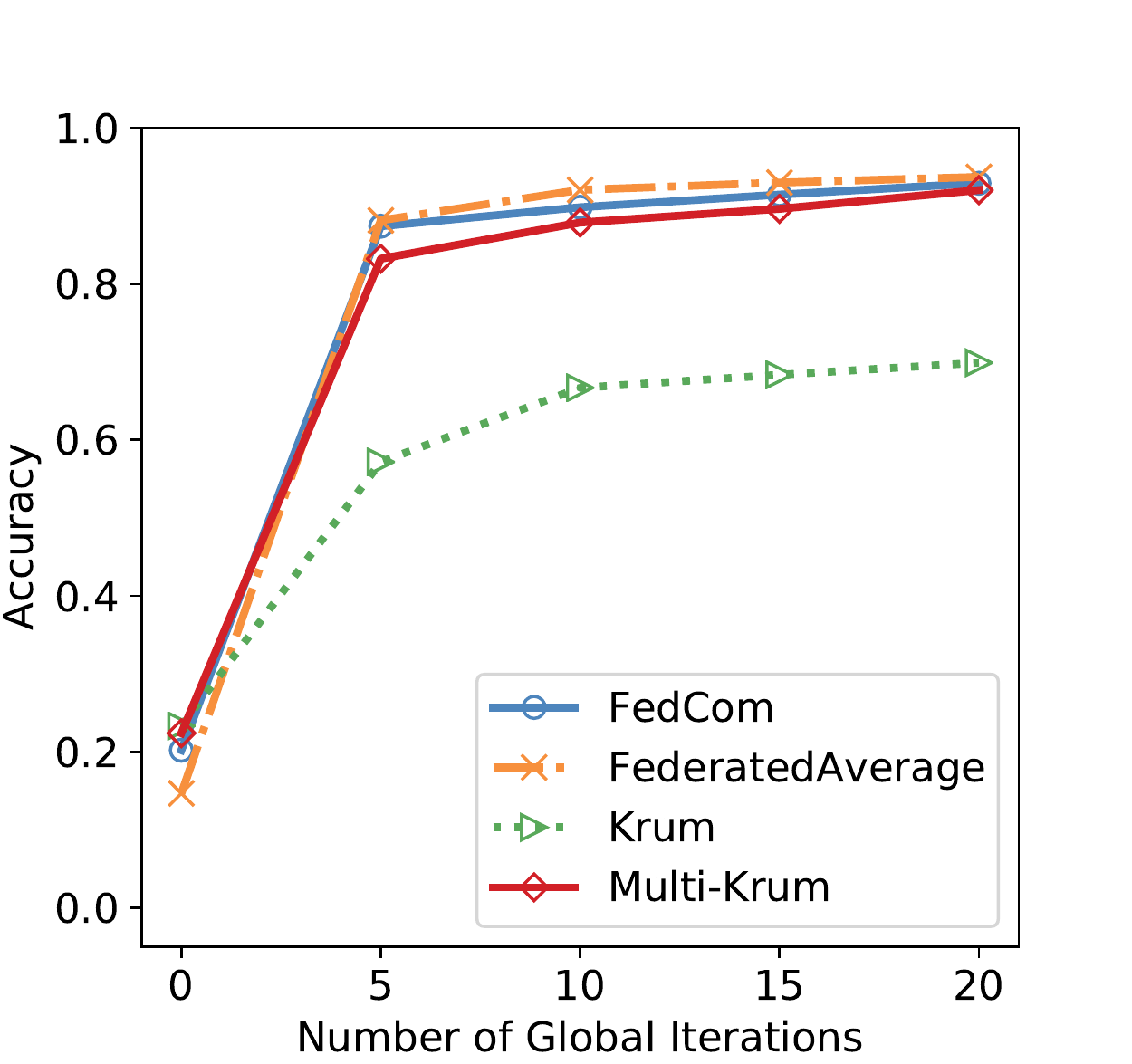}
\centerline{(c) HAR}
\end{minipage}
\begin{minipage}[t]{0.24\textwidth}
\centering
\includegraphics[width=4.8cm]{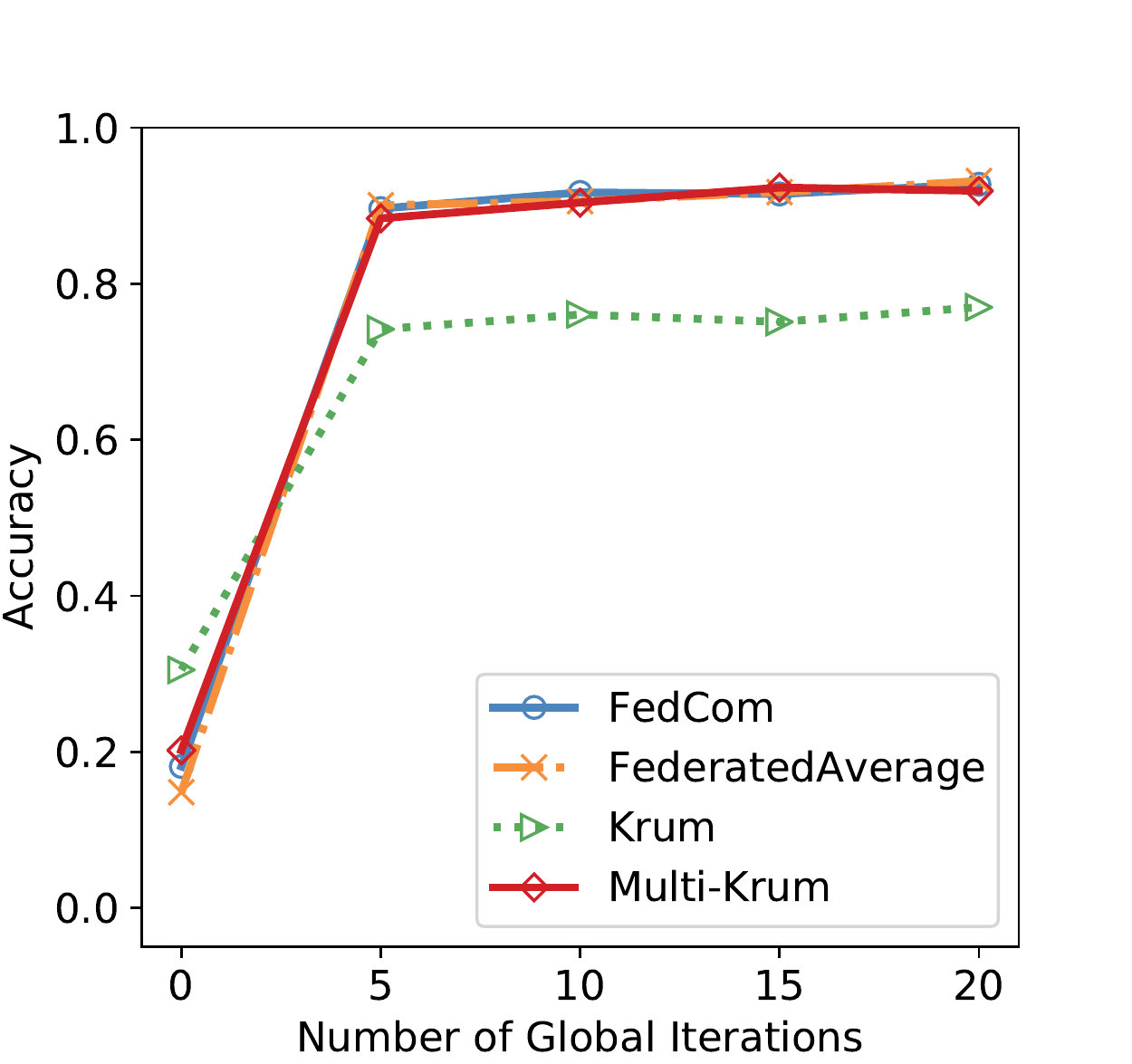}
\centerline{(d) HAR}
\end{minipage}
\caption{Results of model accuracy on different datasets under all aggregation rules. (a) and (c) are results of LR, while (b) and (d) are result of DNN.}
\label{result_noattack}
\end{figure*}

%

We consider a federated learning system which consists of a central server and $20$ truthful workers. The server could employ different aggregation methods for aggregating local model updates to perform global model update. Each independent worker needs to submit the local model update (in each round), the local data size, and the data commitment to the server. We next introduce the datasets, models, baseline Byzantine-robust learning methods, and the attacks considered.

Since data among workers is usually non-IID in practical FL systems, we employ two non-IID datasets for performance evaluation, i.e., Human Activity Recognition (HAR) \cite{APublicDomainDatasetforHumanActivityRecognition} and Non-IID MNIST \cite{LotteryFL:PersonalizedandCommunication-Efficient}. HAR is a dataset of motion data sampled from motion sensors of smart phones, containing 561-feature sample vectors with time and frequency domain variables and 6 different labels (walking, walking upstairs, walking downstairs, sitting, standing and laying). Data in HAR is partitioned by different users, and is naturally Non-IID due to user's different physical conditions. We randomly select 20 users' dataset and assign them to workers as local datasets. MNIST is a dataset for handwritten digit recognition, which consists of 60000 training samples and 10000 testing samples, and each samples is a 28*28 gray-level image. We follow the method in LotteryFL \cite{LotteryFL:PersonalizedandCommunication-Efficient} to generate a non-IID data partition using the standard MNIST dataset. Specifically, we choose "n-class \& unbalanced" setting: each partition contains 10-class training data, and these classes obtain biased feature across the partitions. The volume of each class is unbalanced, and samples are insufficient. Then, we randomly choose 20 local datasets and assign them to workers. Note that the non-IID data distribution in HAR is due to the slightly different physical conditions of volunteers, and thus the non-IID degree is relatively low. In contrast, the processed MNIST data presents a relatively high level of non-IID, as each worker possessed a data partition with biased class distribution or unbalance.


For the machine learning models used in the experiments, we consider both convex and non-convex models. Specifically, we employ a multi-class logistic regression and a deep neural network (DNN). For DNN, we set a hidden layer of 150 neurons and a output layer with Sigmoid function. We adopt Adam with default hyperparameters as optimizer for LR and DNN. It should be noted that our experiment should be focused on performance of other baseline aggregation rules and FedCom facing different poisoning attacks, rather than the best model or hyperparameters for HAR or MNIST. Hence, we do not have to use models with excessive complexity.

In addition to the proposed FedCom, we also implement several baseline aggregation rules for performance comparison, including the vanilla \texttt{FedAverage} as a benchmark and two representative Byzantine-robust FL schemes.

\begin{figure*}[t]
\centering
\begin{minipage}[t]{0.24\textwidth}
\centering
\includegraphics[width=4.8cm]{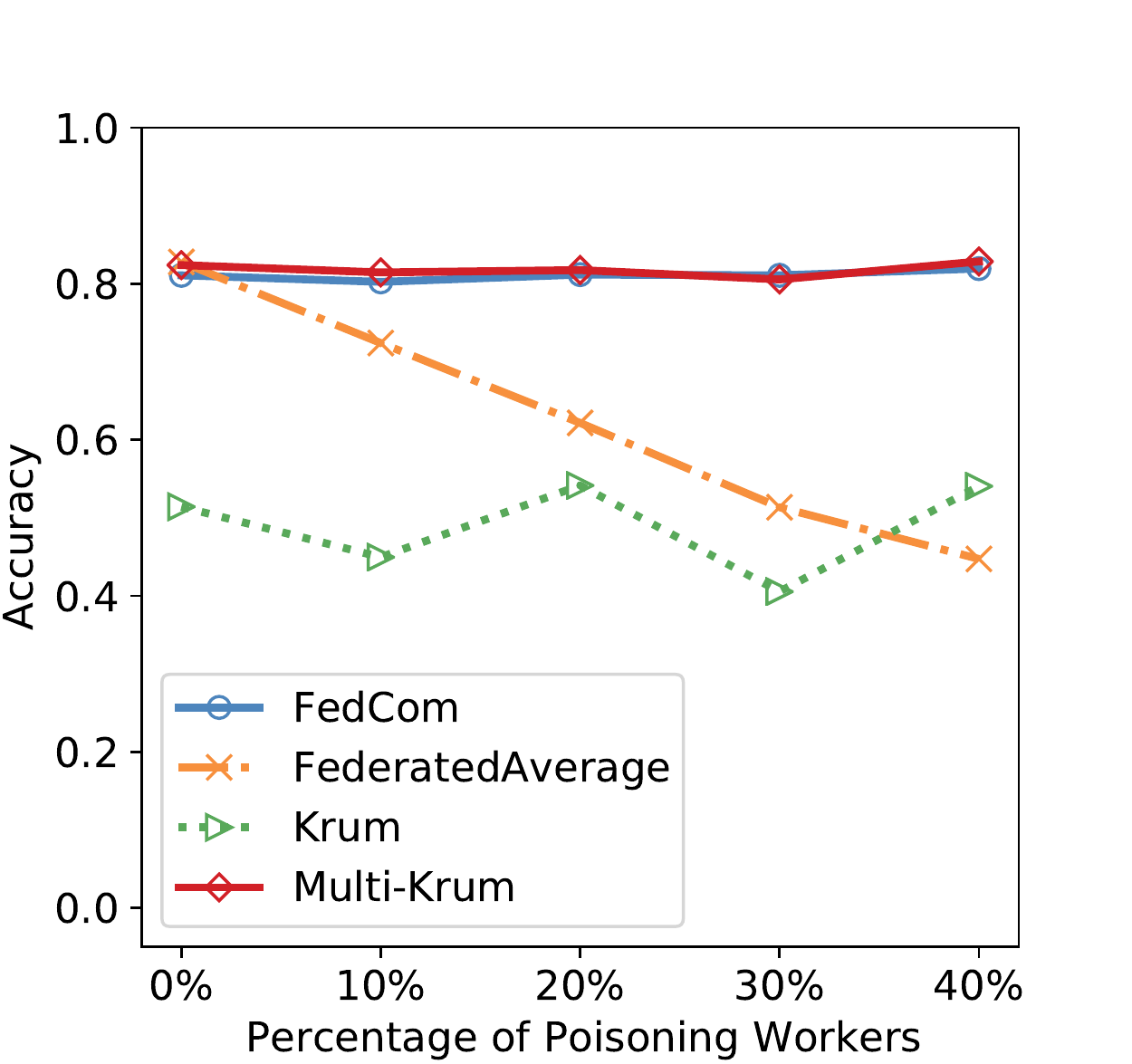}
\centerline{(a) MNIST}
\end{minipage}
\begin{minipage}[t]{0.24\textwidth}
\centering
\includegraphics[width=4.8cm]{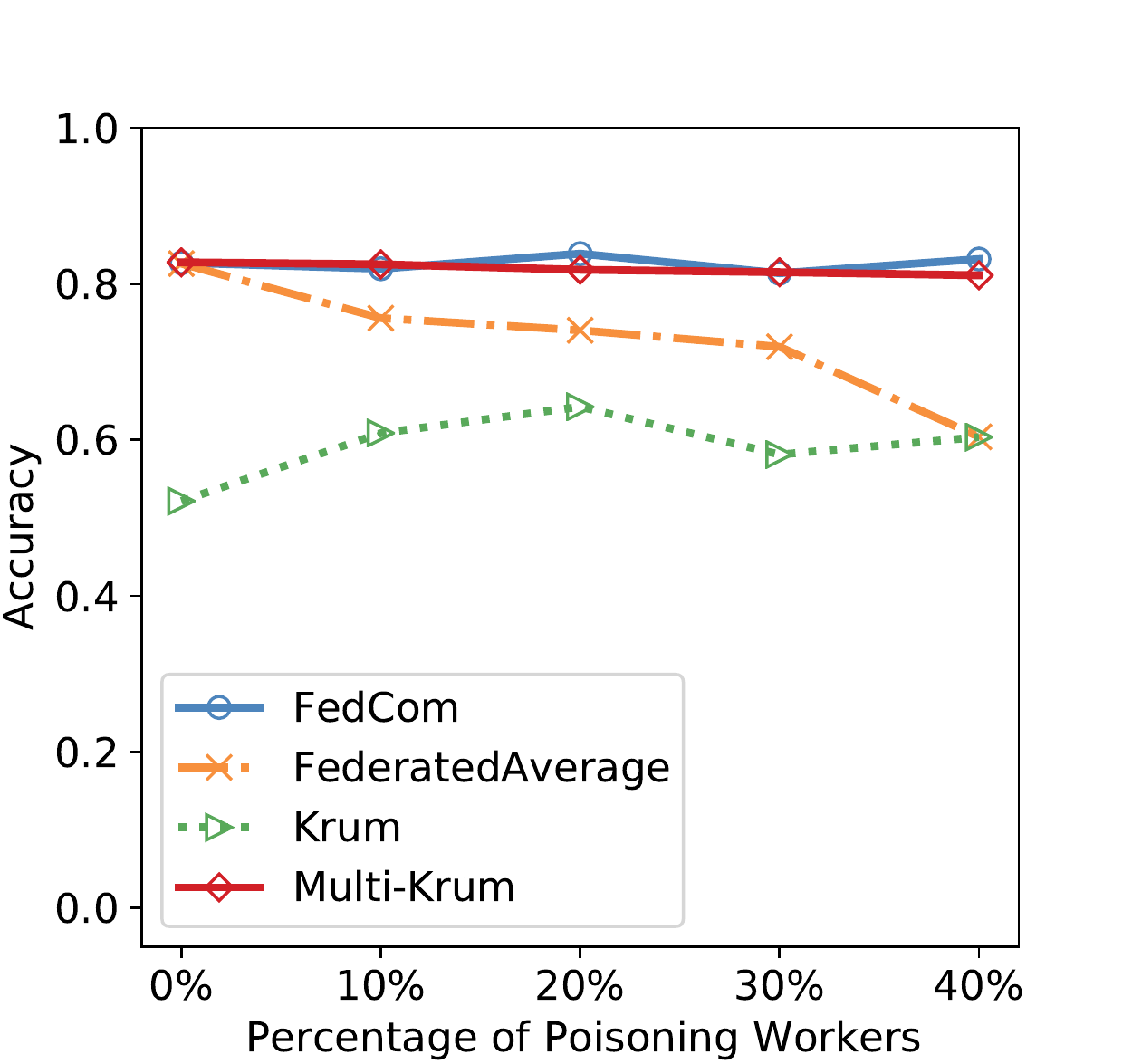}
\centerline{(b) MNIST}
\end{minipage}
\begin{minipage}[t]{0.24\textwidth}
\centering
\includegraphics[width=4.8cm]{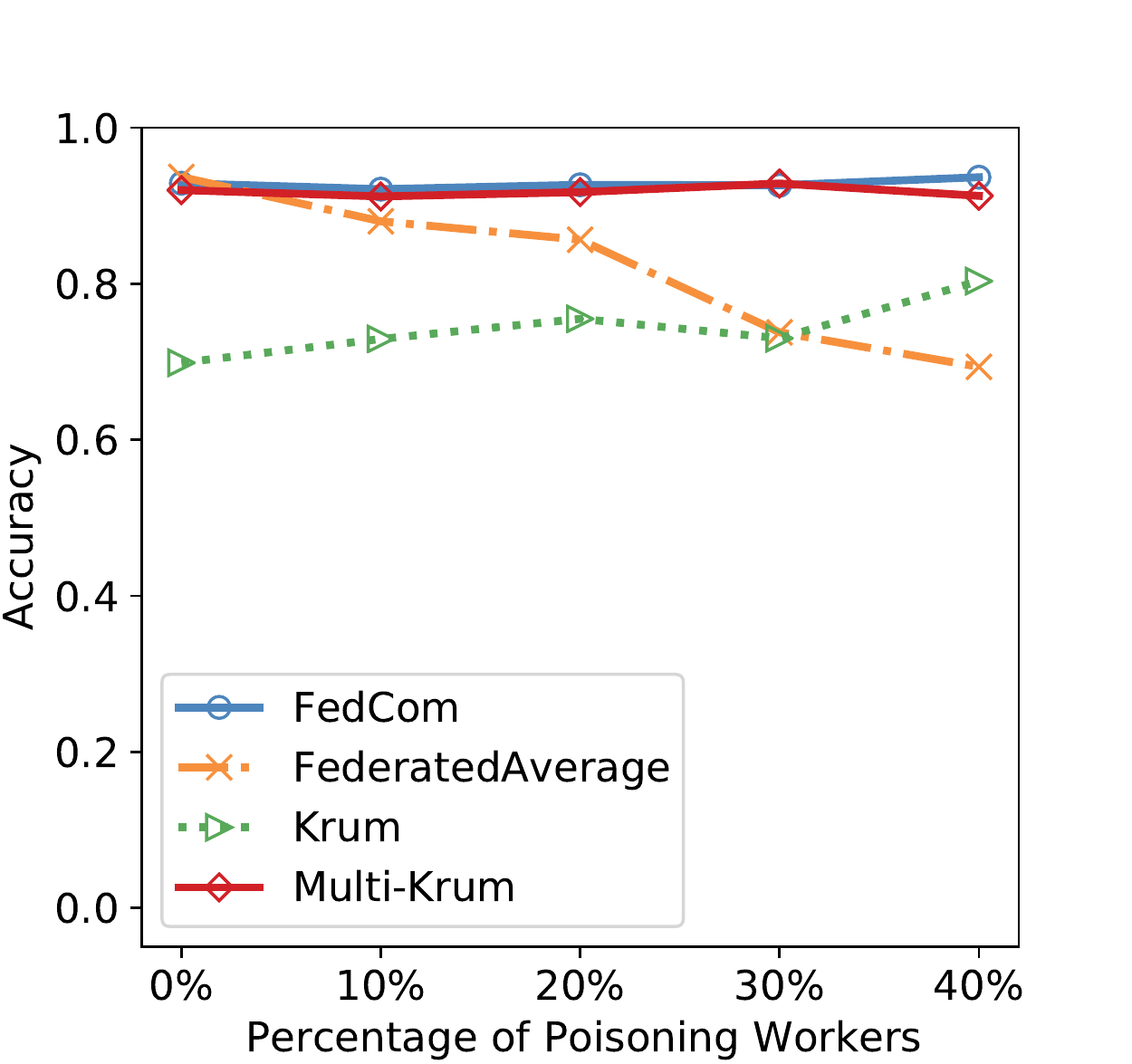}
\centerline{(c) HAR}
\end{minipage}
\begin{minipage}[t]{0.24\textwidth}
\centering
\includegraphics[width=4.8cm]{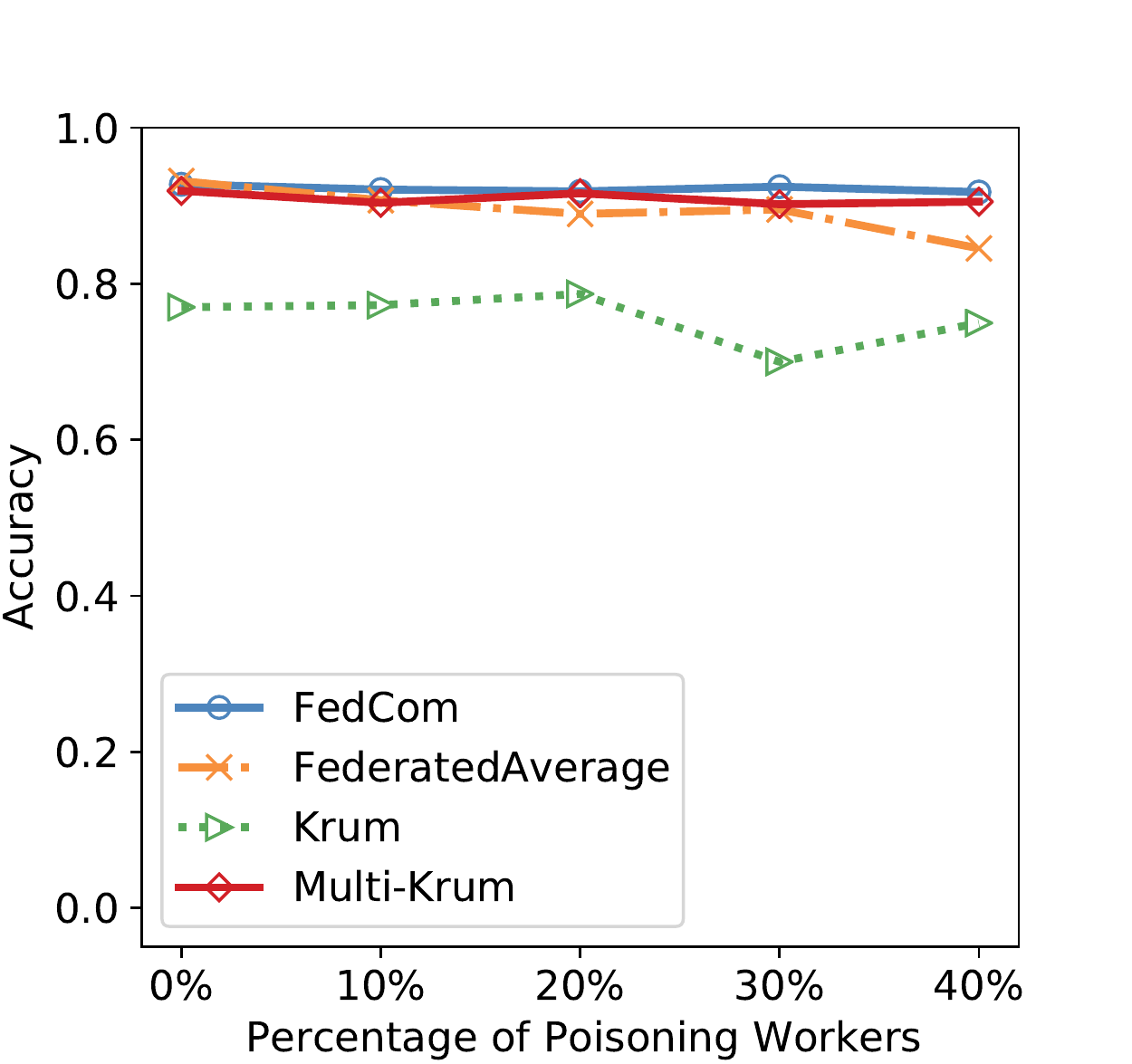}
\centerline{(d) HAR}
\end{minipage}
\caption{Results of model accuracy on benign datasets under Gaussian attack. (a) and (c) are results of LR, while (b) and (d) are result of DNN.}
\label{result_gaussian}
\end{figure*}

\begin{itemize}
  \item \textbf{\texttt{FedAverage}}: the classic aggregation rule of FL, global model is aggregated by averaging local models with weights, in which weights are the ratio of local dataset's size and global dataset's size.
  \item \textbf{\texttt{Krum}}: representing a series of DSA schemes. \texttt{Krum} assumes that benign local models obtain closer Euclidian distances, and calculate scores for each local model: find $n-f-1$ closest local models to current local model, and calculate the sum of Euclidian distances between these models and current local model as score. \texttt{Krum} selects the local model with lowest score as global model.
  \item \textbf{\emph{Multi-}\texttt{Krum}}: constructing a combination of \texttt{FedAverage} and \texttt{Krum} to keep the generality of \texttt{FedAverage} on Non-IID local datasets while introducing the robustness of \texttt{Krum}. \emph{Multi-}\texttt{Krum} adopt the same scoring rule as \texttt{Krum}, but take $n-f-1$ local models with lowest score, and aggregate these local models via \texttt{FedAverage}.
\end{itemize}

\texttt{FedAverage} contains no Byzantine robustness, hence is taken as a baseline to verify the effectiveness of poisoning attacks. Many schemes inherit the core idea of \texttt{Krum} and \emph{Multi-}\texttt{Krum}, which is eliminating local models or dimensions that are far away from the majority. Hence, \texttt{Krum} and \emph{Multi-}\texttt{Krum} could be representative baselines of DSA schemes. It should be noticed that we consider schemes of CSA are difficult to be implemented in FL, hence such schemes are not in the scope of our evaluation.

We implemented two types of LDP and two types of LMP to test performance of four aggregation rules.
\begin{itemize}
  \item \textbf{Label flipping}: a naive LDP by simply mislabelling training samples to change the distribution of dataset. Concretely, we adopt the strategy of rotation. For example, in MNIST, we label digit "1" as "2", digit "2" as "3", and so on.
  \item \textbf{Back-gradient} \cite{TowardsPoisoningofDeepLearningAlgorithmswithBack-gradient}: an advanced LDP by solving a reversed optimizing problem on training set. Concretely, samples in training set is adjusted along the gradient that a well-trained model's loss on validation set could increase, and finally obtain a training set that maximize the model's loss on validation set.
  \item \textbf{Gaussian attack}: a naive LMP by submitting Gaussian noise as a local model, attempting to prevent global model's convergence.
  \item \textbf{Krum attack} \cite{LocalModelPoisoningAttacksto}: an advanced LMP aimed at \texttt{Krum} or similar aggregation rules. Such attack exploits the vulnerability of \texttt{Krum} and similar aggregation rules, and attemp to construct a poisoning model that is the closest to other local models, but obtain reverse direction to correct global gradient. Such poisoning local models could always been chosen by \texttt{Krum} to become global model, or participate in global model's aggregation.
\end{itemize}

Concretely, we implement these attacks on several workers locally, and inject these Byzantine workers into FL system. For every aggregation rule, we set the percentage of Byzantine workers from 0\% (no attack) to 40\%. For LMP, we observe global model's accuracy on benign validation set. For LDP, we observe global model's accuracies on both benign validation set and poisoned dataset, as LDP could also be targeted poisoning attack, and sometimes the attack effect could only be reflected to the increase accuracy of global model on poisoned dataset.

Besides, we need to consider a comprehensive threat model, hence, for FedCom, we specifically designed two extra attacking method: \emph{honest commitment} (HC) and \emph{fake commitment} (FC). This is because poisoning attackers could choose to construct data commitment honestly using poisoning local datasets, or choose to construct a fake data commitment using a benign local dataset. Such a variate is beyond control of FedCom, as a result, it is necessary to separately discuss such two attacking methods. Noting that LMP attackers do not obtain related data distribution of poisoning local models they craft, hence we consider all LMP as FC attacks.

\subsection{Basic Evaluation of Aggregation Rules}

In the beginning of our evaluation, we should evaluate the effectiveness of aggregation rules, to eliminate the possibility of unexpected factors like implementation error. We perform FL task separately using LR and DNN on two datasets under four aggregation rules, and workers are all benign. Figure \ref{result_noattack} shows the accuracies of global models after several global iterations, and we will discuss about such results from aspects below.

\textbf{Effectiveness of aggregation rules}: the accuracy of global model will always converge to a certain upper bound even switching different aggregation rules, which shows that four implemented aggregation rules are all effective to enhance the accuracy of global models in an FL task.

\textbf{Impact of Non-IID}: note that the Non-IID feature of two datasets negatively impacts the accuracy of global models. Local datasets of workers in our evaluation are randomly chosen and assigned to workers from a batch of local datasets, and neither of the global data distributions of two datasets are intact. Hence, the accuracy of global models on testing sets may not reach to a definite favourable level. However, it has nothing to do with comparing robustness between four aggregation rules. The results of all aggregation rules meet the expectation. Firstly, for \texttt{Krum}, the Non-IID local datasets will significantly reduce the accuracy of global model, as global model under \texttt{Krum} is the closest local model to other models in the majority group, and a global model which overfits a certain local dataset will fit other local datasets well, hence will not obtain favourable performance on global validation set. By contrast, FedCom, \texttt{FedAverage} and \emph{Multi-}\texttt{Krum} maintain or inherit the weighted average mechanism in \texttt{FedAverage}, hence obtain better performance on Non-IID local datasets. Secondly, for HAR and MNIST that both of them obtain deletion of global data distribution, global model's accuracy on HAR could relatively higher than that on MNIST, and this is because the Non-IID degree of HAR is smaller than that of MNIST, which also means that the bias degree of local models is smaller, hence the global model's accuracy is better.

Through results in Figure \ref{result_noattack}, we could conclude that \textbf{FedCom could obtain similar performance, as well as tolerance of Non-IID local datasets, to aggregation rules like \texttt{FedAverage} or \emph{Multi-}\texttt{Krum}}.  For convenience, we take final accuracy of such results as ideal accuracy under related aggregation rules, and compare such results to those under different poisoning attacks to show the robustness of aggregation rules when facing these attacks.

\subsection{Results for Naive Poisoning Attacks}

\begin{figure*}[t]
\centering
\begin{minipage}[t]{0.24\textwidth}
\centering
\includegraphics[width=4.8cm]{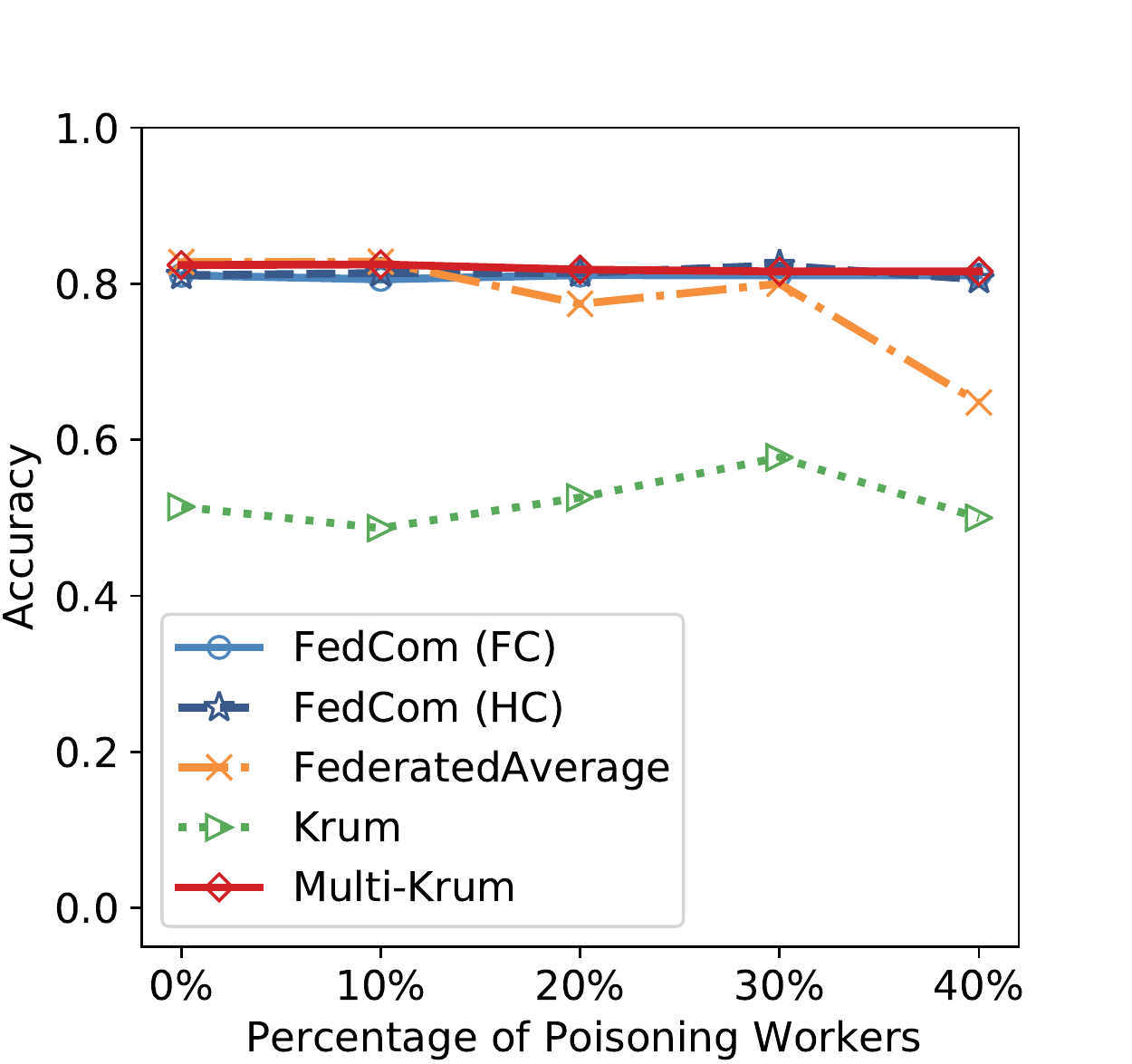}
\centerline{(a) MNIST}
\end{minipage}
\begin{minipage}[t]{0.24\textwidth}
\centering
\includegraphics[width=4.8cm]{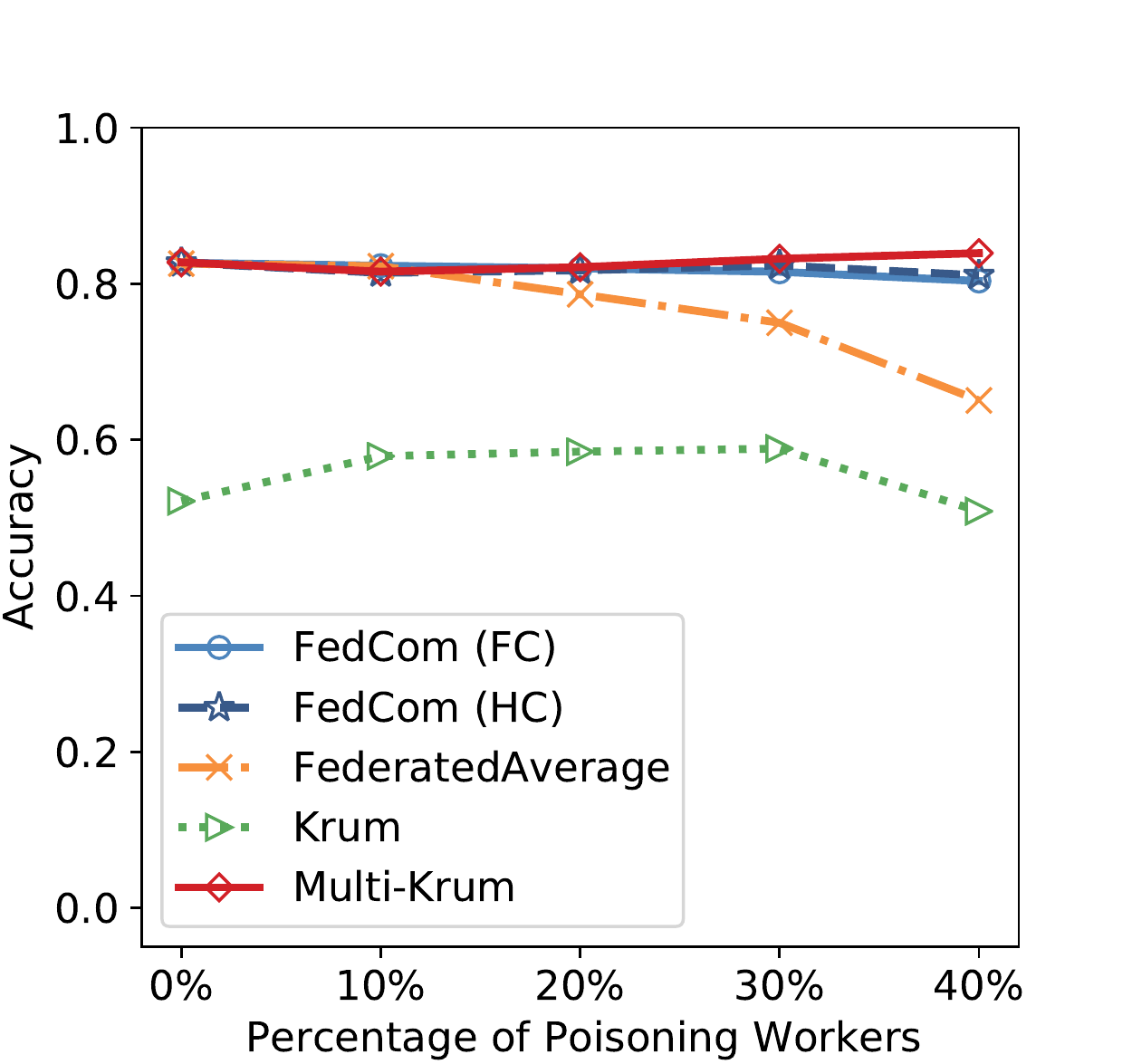}
\centerline{(b) MNIST}
\end{minipage}
\begin{minipage}[t]{0.24\textwidth}
\centering
\includegraphics[width=4.8cm]{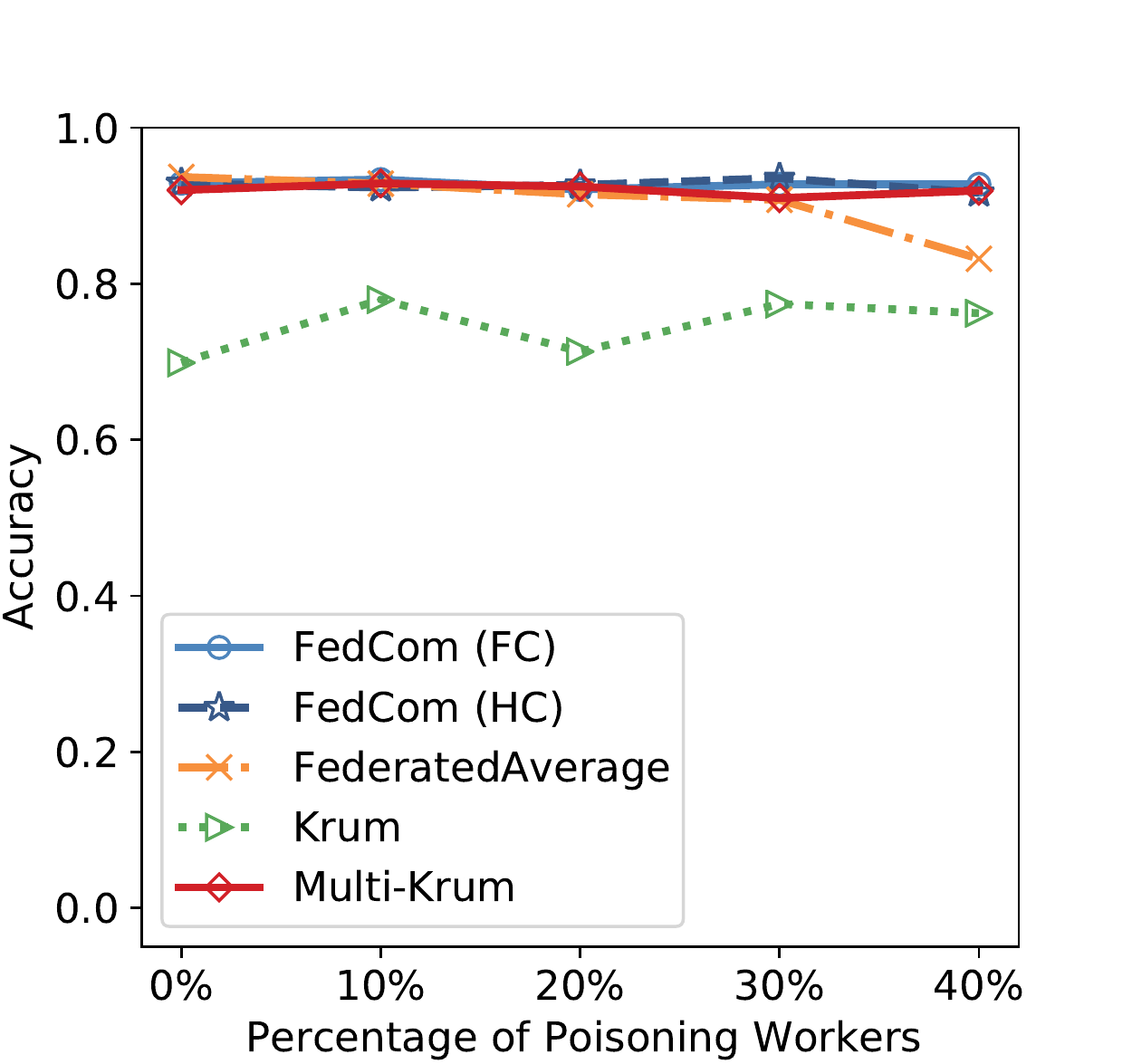}
\centerline{(c) HAR}
\end{minipage}
\begin{minipage}[t]{0.24\textwidth}
\centering
\includegraphics[width=4.8cm]{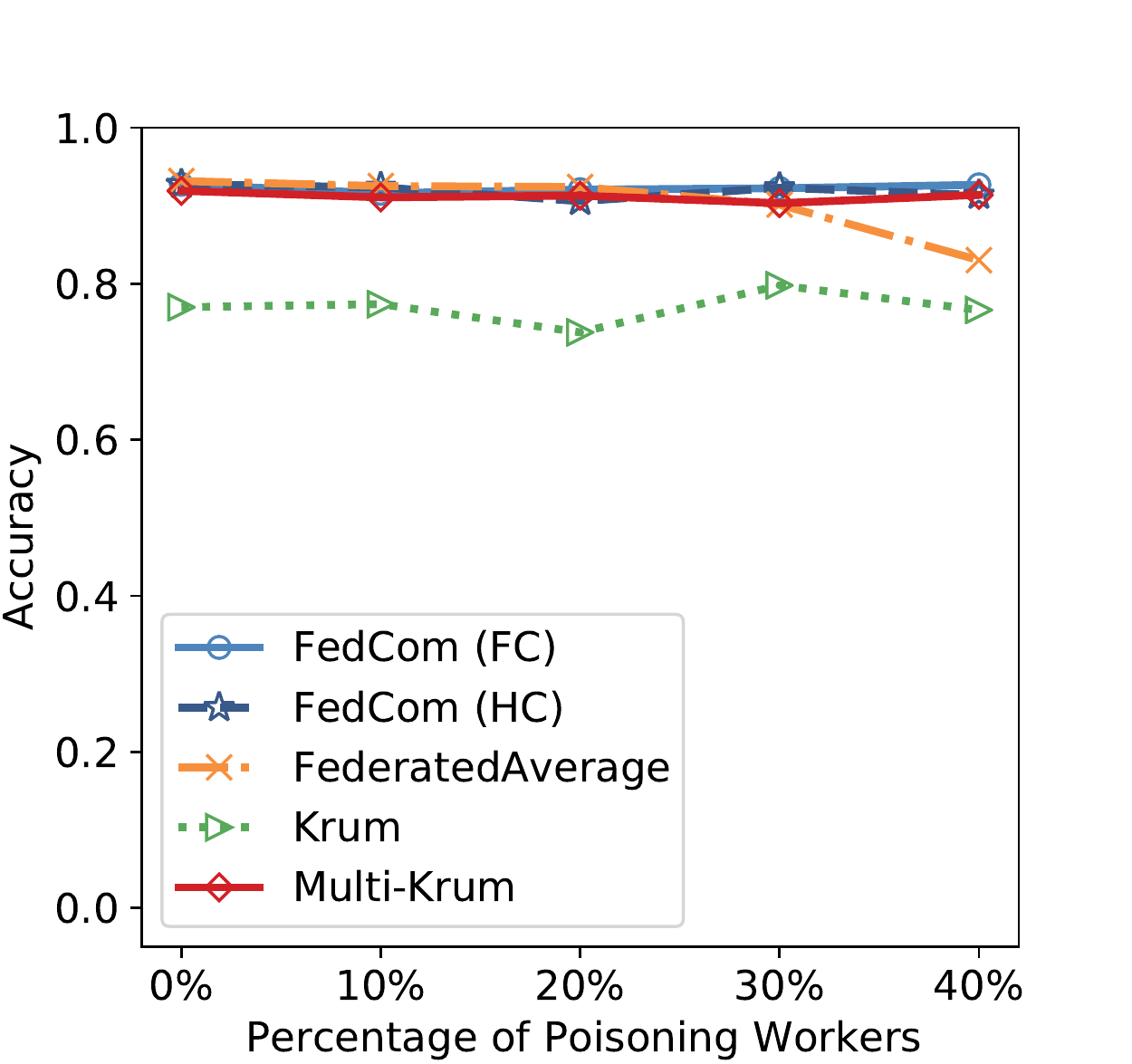}
\centerline{(d) HAR}
\end{minipage}
\begin{minipage}[t]{0.24\textwidth}
\centering
\includegraphics[width=4.8cm]{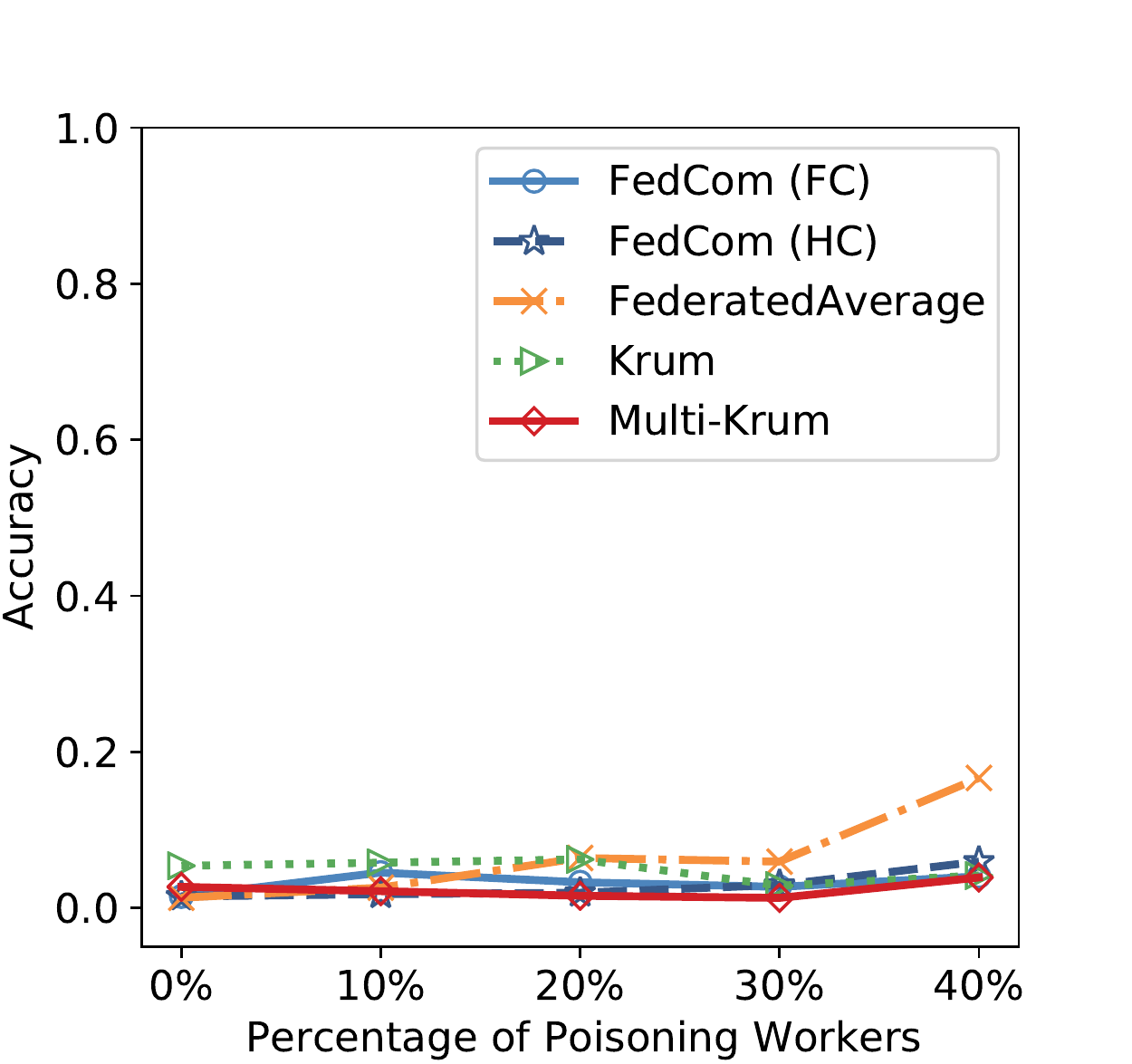}
\centerline{(e) MNIST}
\end{minipage}
\begin{minipage}[t]{0.24\textwidth}
\centering
\includegraphics[width=4.8cm]{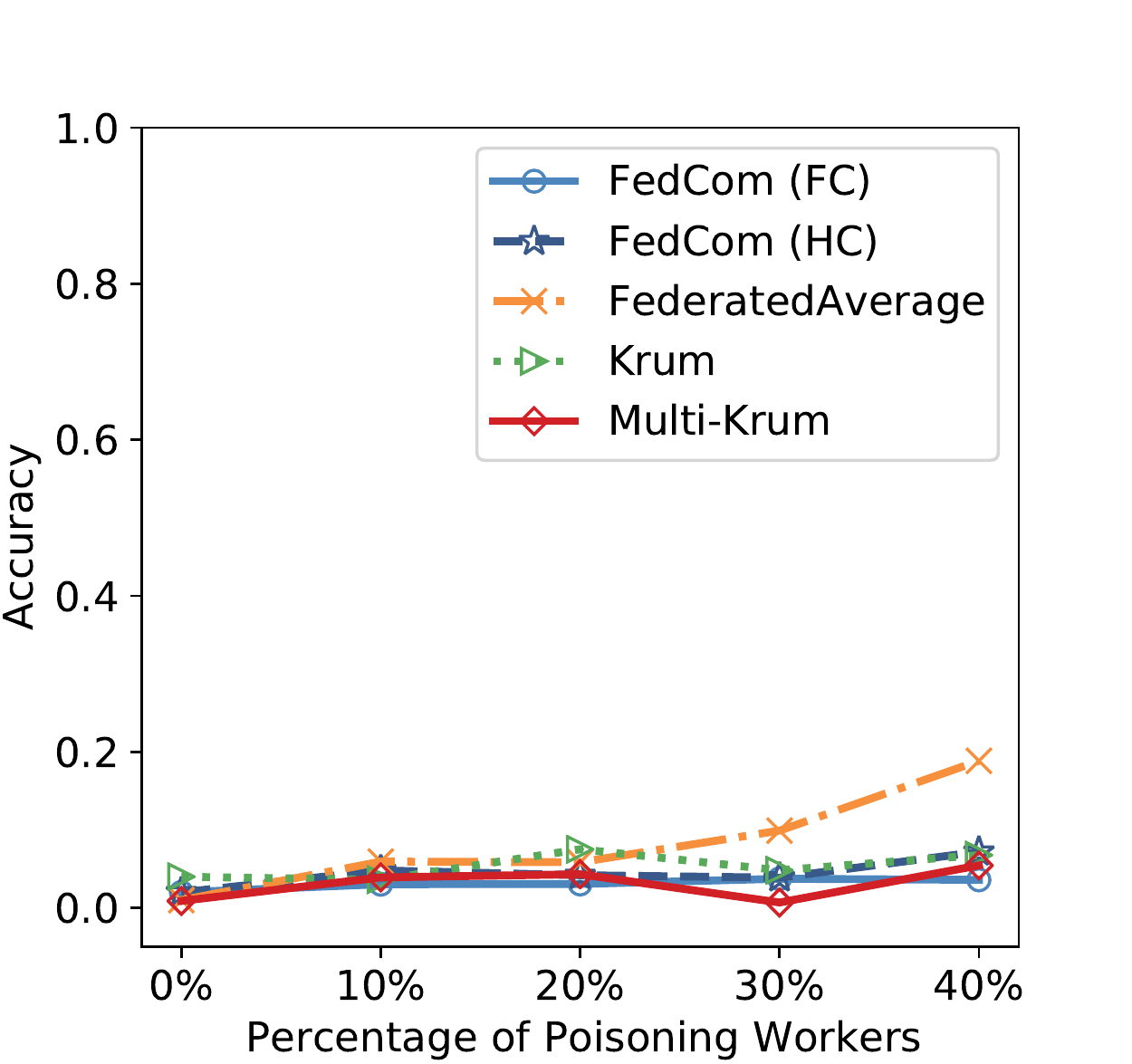}
\centerline{(f) MNIST}
\end{minipage}
\begin{minipage}[t]{0.24\textwidth}
\centering
\includegraphics[width=4.8cm]{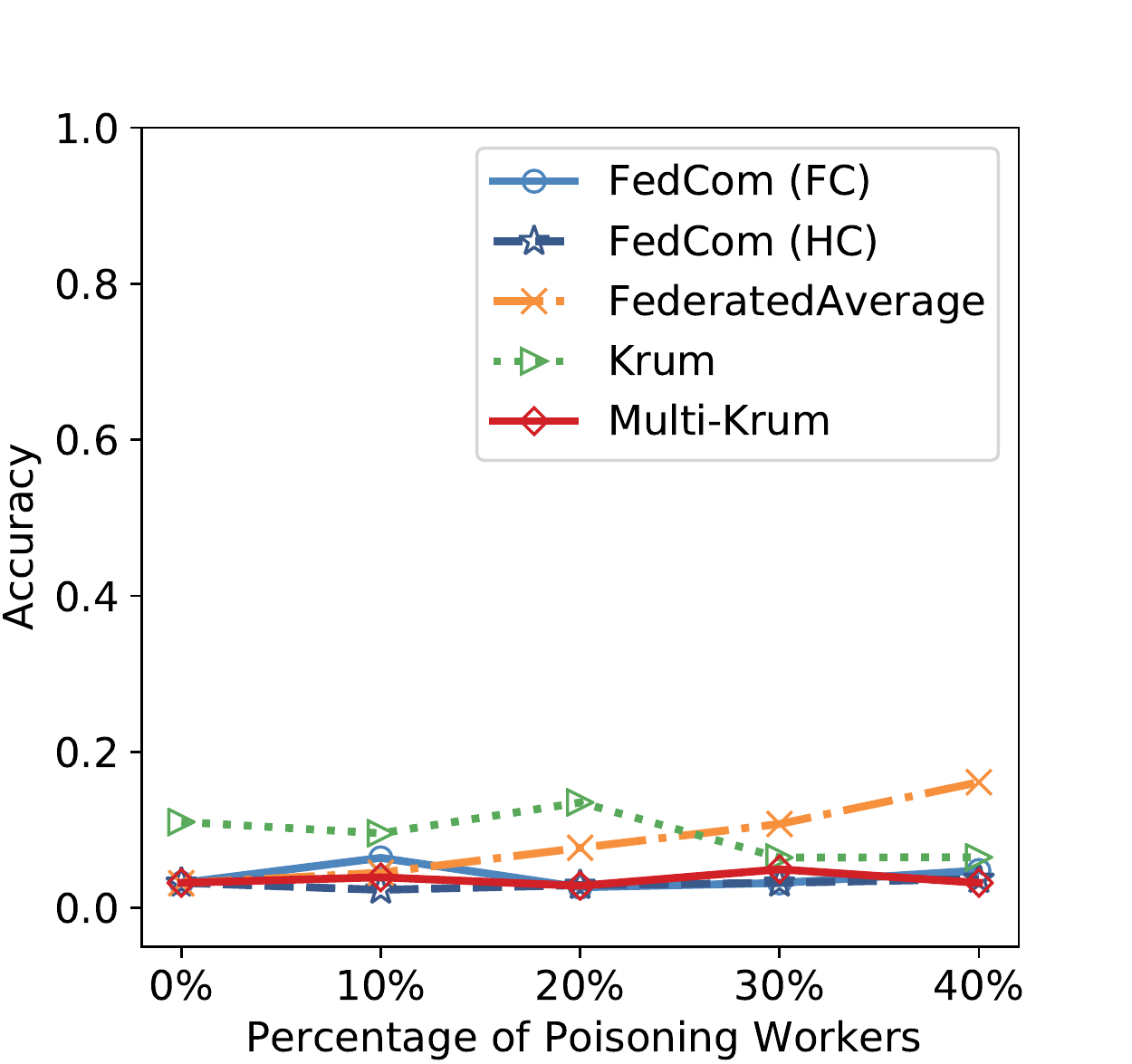}
\centerline{(g) HAR}
\end{minipage}
\begin{minipage}[t]{0.24\textwidth}
\centering
\includegraphics[width=4.8cm]{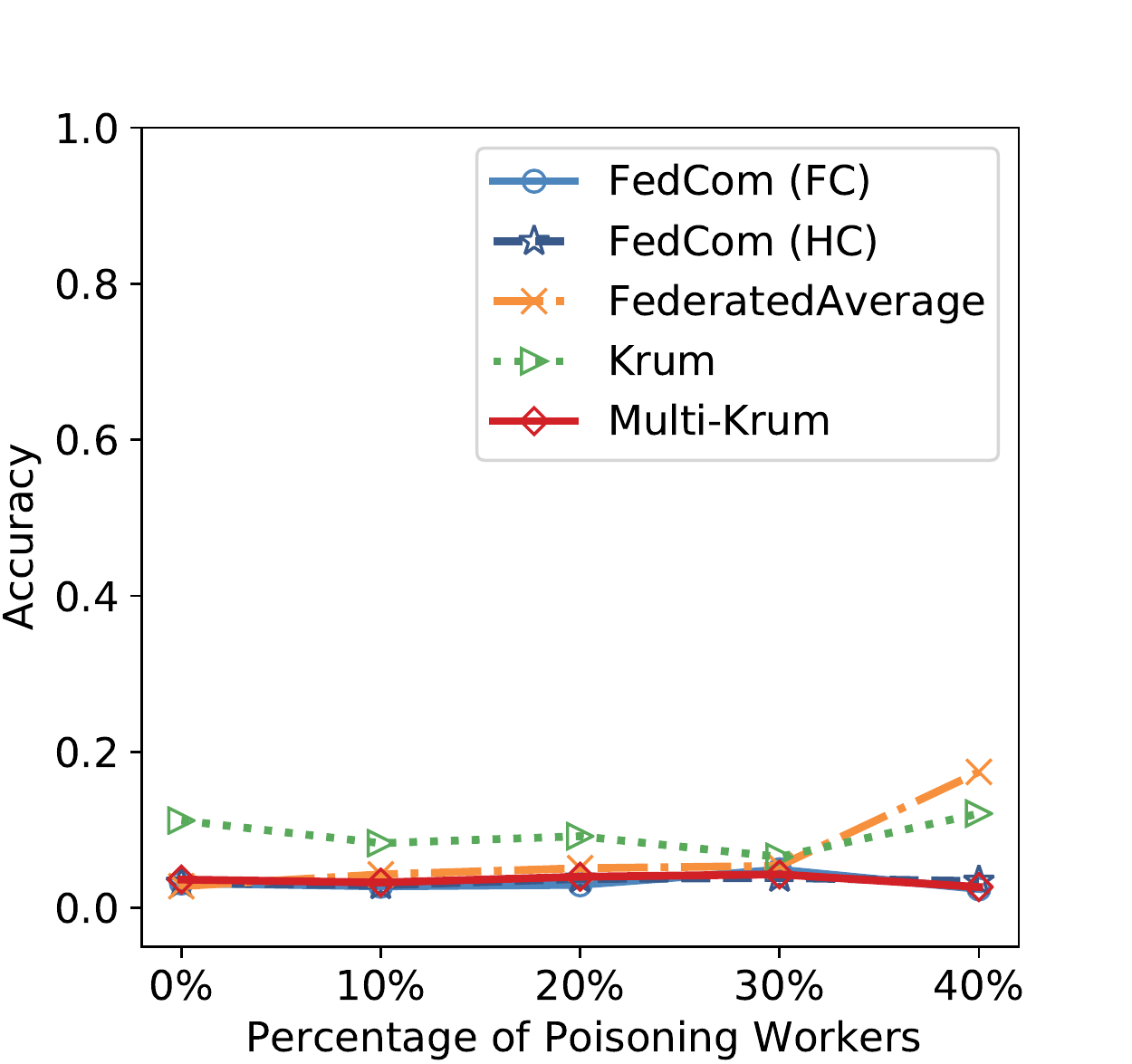}
\centerline{(h) HAR}
\end{minipage}
\caption{Results of model accuracy on benign and poisoned datasets under Label-flipping attack. (a) and (c) are results of LR on benign datasets, while (b) and (d) are result of DNN on benign datasets. (e) and (g) are results of LR on label-flipped datasets, while (f) and (h) are results of DNN on label-flipped datasets.}
\label{result_labelflip}
\end{figure*}

We firstly implement the compete between naive poisoning attacks and four aggregation rules to preliminarily test their robustness against poisoning attacks. We adopt Label-flipping attack and Gaussian attack as LDP and LMP. For Gaussian attack, we collect accuracy of global models on benign validation set, and for Label-flipping attack, we collect accuracy of global models on both benign validation set and label-flipped dataset. Figure \ref{result_gaussian} shows the results of Gaussian attack and Figure \ref{result_labelflip} shows the results of Label-flipping attack. We will discuss about such results from aspects below.

\textbf{Robustness against poisoning attack}: we could summarize from the results that even if facing naive attacks, \texttt{FedAverage} is unable to effectively defend. Under Gaussian attack, global models obtain decay of accuracy on benign validation sets. under Label-filpping attack, the accuracy on label-flipped datasets rises up, and accuracy on benign validation set decreases. Such phenomenon meets our expectation, as \texttt{FedAverage} is proved to obtain no Byzantine robustness. By contrast, other three aggregation rules all perform various degree of robustness. Facing Gaussian and Label-flipping attack, global models under FedCom and \emph{Multi-}\texttt{Krum} obtain the same level of accuracy as there is no attack. Global models under \texttt{Krum} obtain unstable performance facing two attacks, but we consider such phenomenon rational, as \texttt{Krum} may select different local model as global model under different percentage of poisoning model, and different local model performs differently on benign validation set due to Non-IID local datasets. Nevertheless, the performance of global models under \texttt{Krum} is never worse than that when there is no attack, and this could reflect the robustness of \texttt{Krum}. When facing Label-flipping attack, three robust aggregation rules could constrain the accuracy of global models on label-flipped dataset to a certain low rate, which means that local models trained by label-flipped local datasets are almost isolated from global model's aggregation. It is shown that the naive poisoning attacks we implemented are correct, and FedCom, \emph{Multi-}\texttt{Krum} and \texttt{Krum} obtain different levels of poisoning attack robustness.

\textbf{Impact of percentage of poisoning workers}: obviously, global models under \texttt{FedAverage} obtain worse performance with larger percentage of poisoning workers facing both Gaussian attack and Label-flipping attack. This indicates that both two attacks obtain better effectiveness with more poisoning workers in a FL system. However, at any percentage of poisoning workers in our evaluation, the performance of global model under FedCom, \texttt{Krum} or \emph{Multi-}\texttt{Krum} is not worse than that when there is no attack. This indicates that such three aggregation rules obtain safety threshold of over 40\%.

\textbf{Impact of Non-IID}: we need to note that the degree of Non-IID will impact the effectiveness of poisoning attacks. For Gaussian attack, we could observe that the accuracy of global model on MNIST under \texttt{FedAverage} obtains larger percentage of decay than that on HAR. For example, LR under \texttt{FedAverage} facing Gaussian attack obtains decay of accuracy on MNIST at percentage of 45.12\%, and this percentage on HAR is only 27.17\%. This indicates that higher degree of Non-IID makes Gaussian attack more effective. Such phenomenon could be rationally explained: a higher degree of Non-IID of local datasets makes benign local models more dispersive, and the length of benign global gradient aggregated by these benign models is shorterFor poisoning models, a shorter benign global gradient length could make manipulating the direction of final global gradient easier. Such phenomenon also occurs in Label-flipping attack, and explanation above is also appropriate. However, the results also indicated that the higher degree of Non-IID does not impact the robustness of FedCom, \emph{Multi-}\texttt{Krum} or \texttt{Krum}.

Through results above, we could summarize that \textbf{FedCom obtains similar robustness to \emph{Multi-}\texttt{Krum} or \texttt{Krum} while facing naive attacks like Label-flipping attack or Gaussian attack, and FedCom, as well as \emph{Multi-}\texttt{Krum}, obtains well tolerance to Non-IID's impact while retaining robustness}.

\subsection{Results for Advanced Poisoning Attacks}

\begin{figure*}[t]
\centering
\begin{minipage}[t]{0.24\textwidth}
\centering
\includegraphics[width=4.8cm]{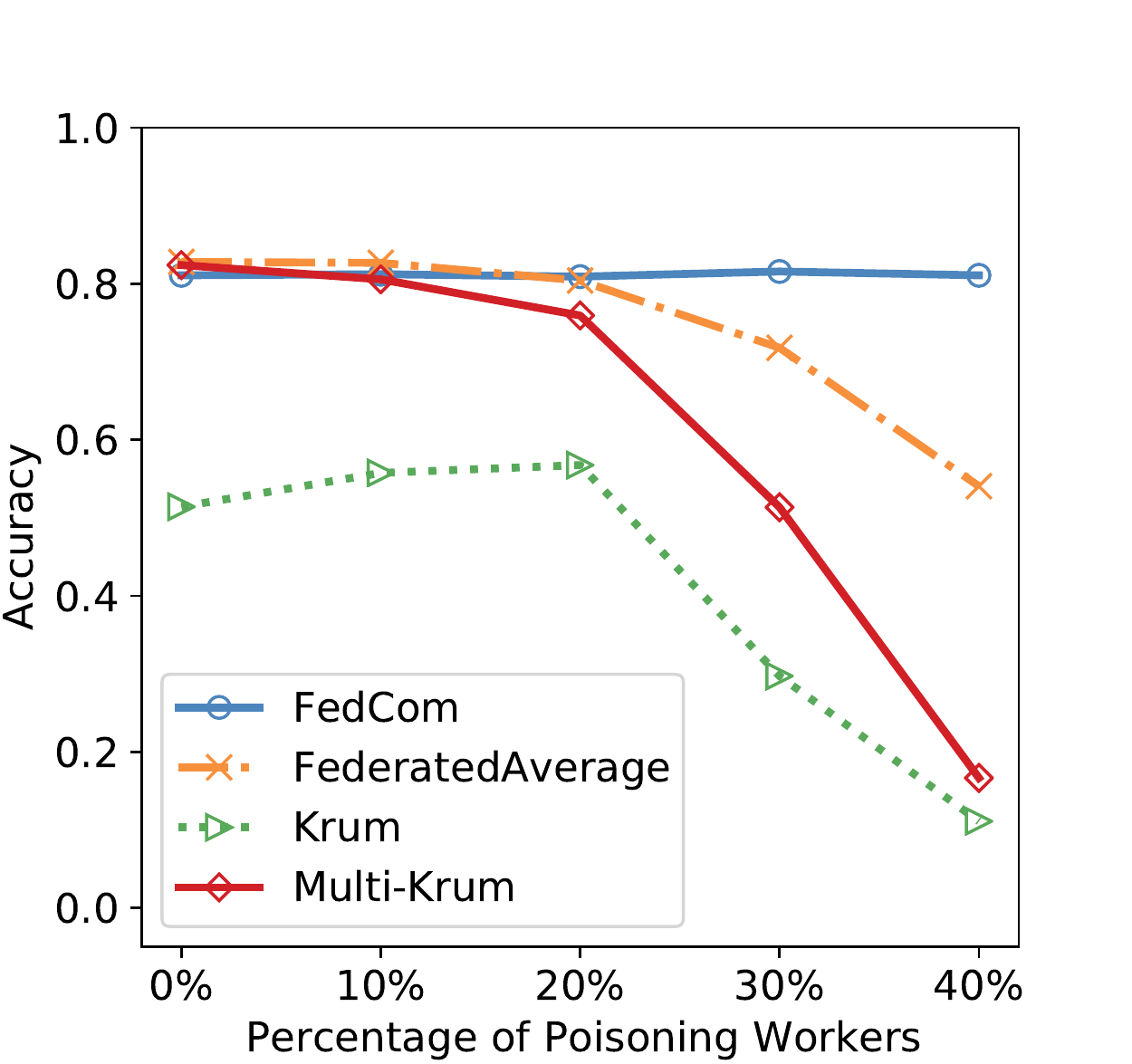}
\centerline{(a) MNIST}
\end{minipage}
\begin{minipage}[t]{0.24\textwidth}
\centering
\includegraphics[width=4.8cm]{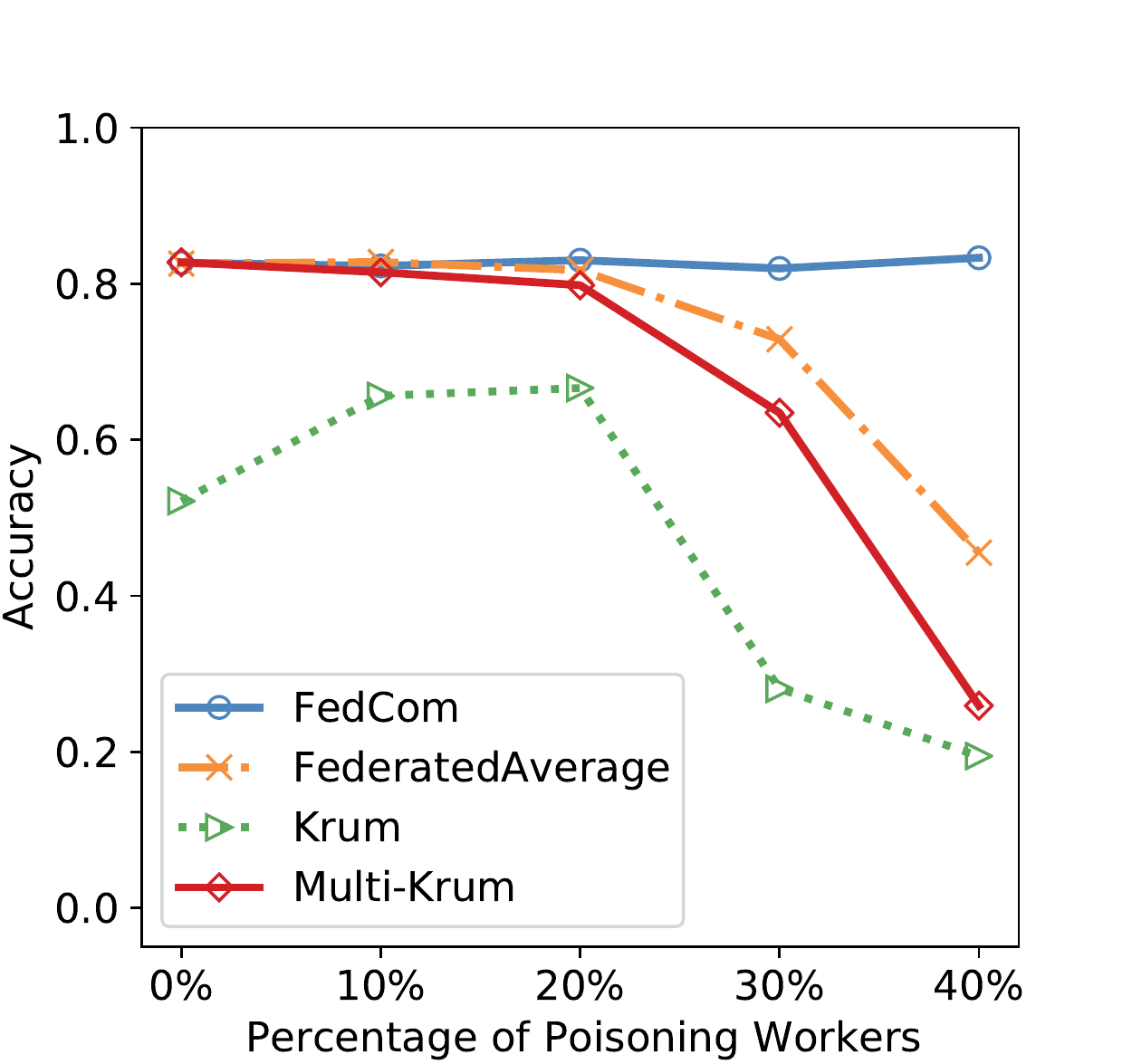}
\centerline{(b) MNIST}
\end{minipage}
\begin{minipage}[t]{0.24\textwidth}
\centering
\includegraphics[width=4.8cm]{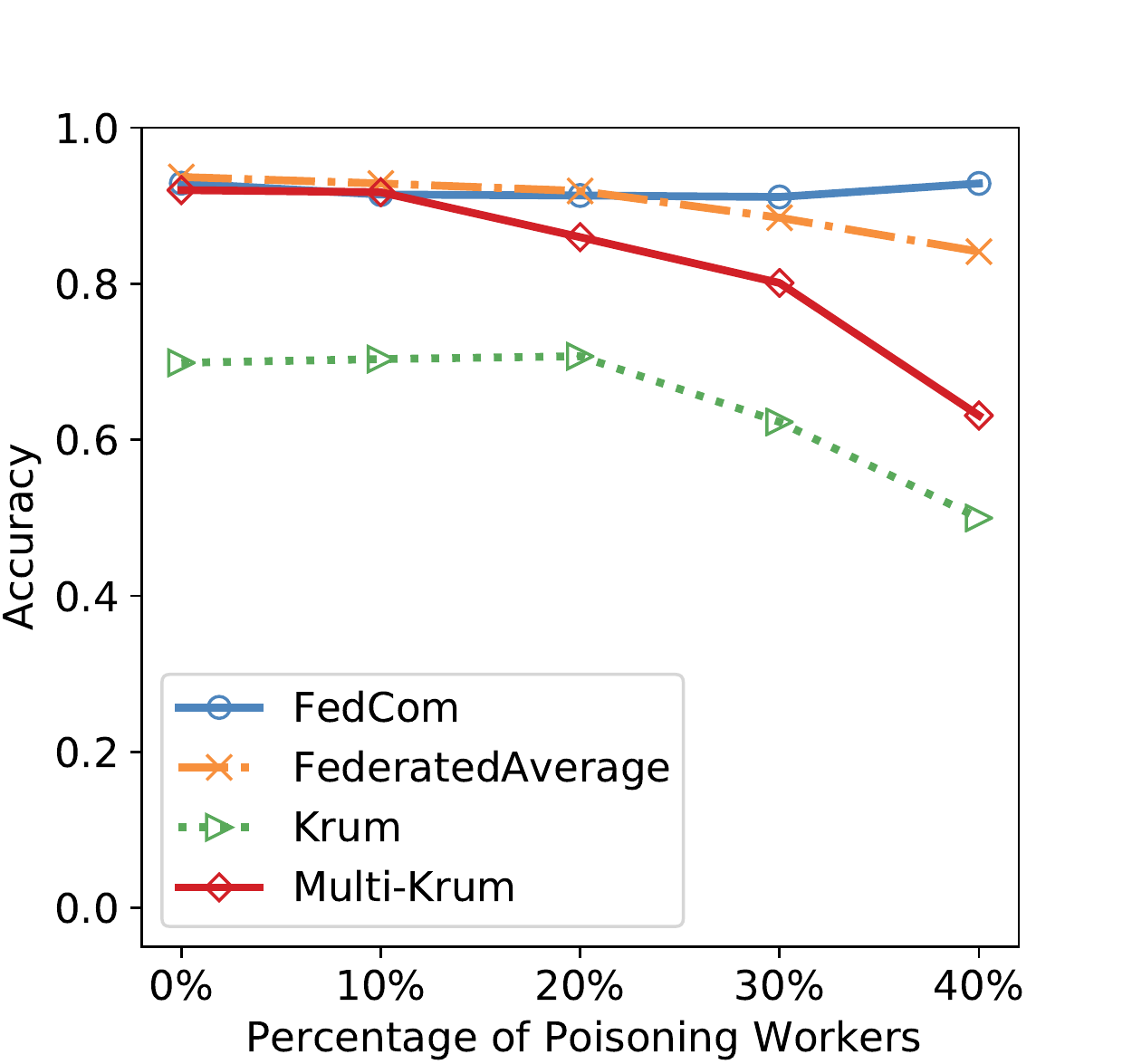}
\centerline{(c) HAR}
\end{minipage}
\begin{minipage}[t]{0.24\textwidth}
\centering
\includegraphics[width=4.8cm]{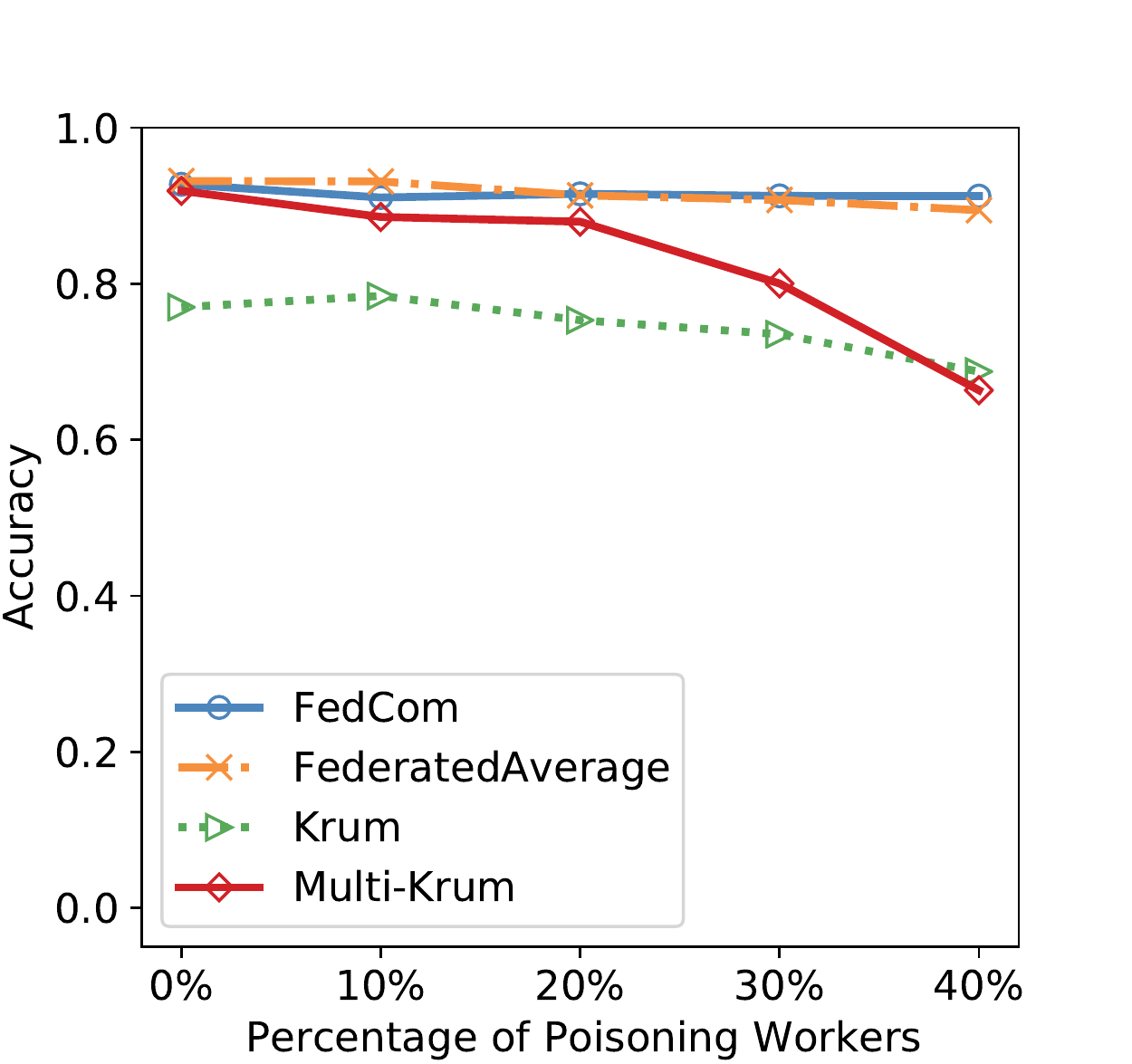}
\centerline{(d) HAR}
\end{minipage}
\caption{Results of model accuracy on benign datasets under Krum attack. (a) and (c) are results of LR, while (b) and (d) are result of DNN.}
\label{result_krumattack}
\end{figure*}

\begin{figure}[t]
\centering
\begin{minipage}[t]{0.23\textwidth}
\centering
\includegraphics[width=4.5cm]{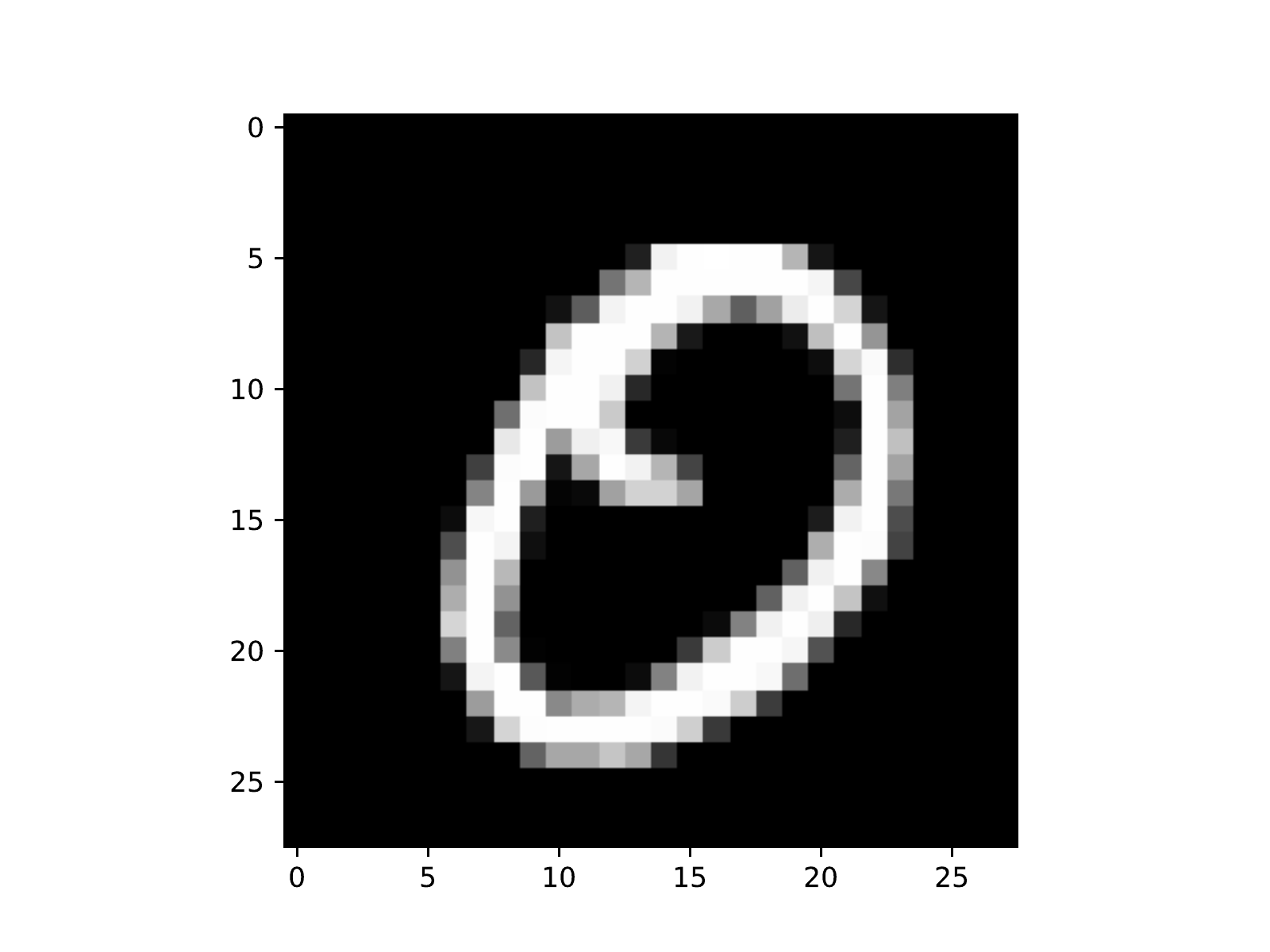}
\centerline{(a) "0" before Back-gradient}
\end{minipage}
\begin{minipage}[t]{0.23\textwidth}
\centering
\includegraphics[width=4.5cm]{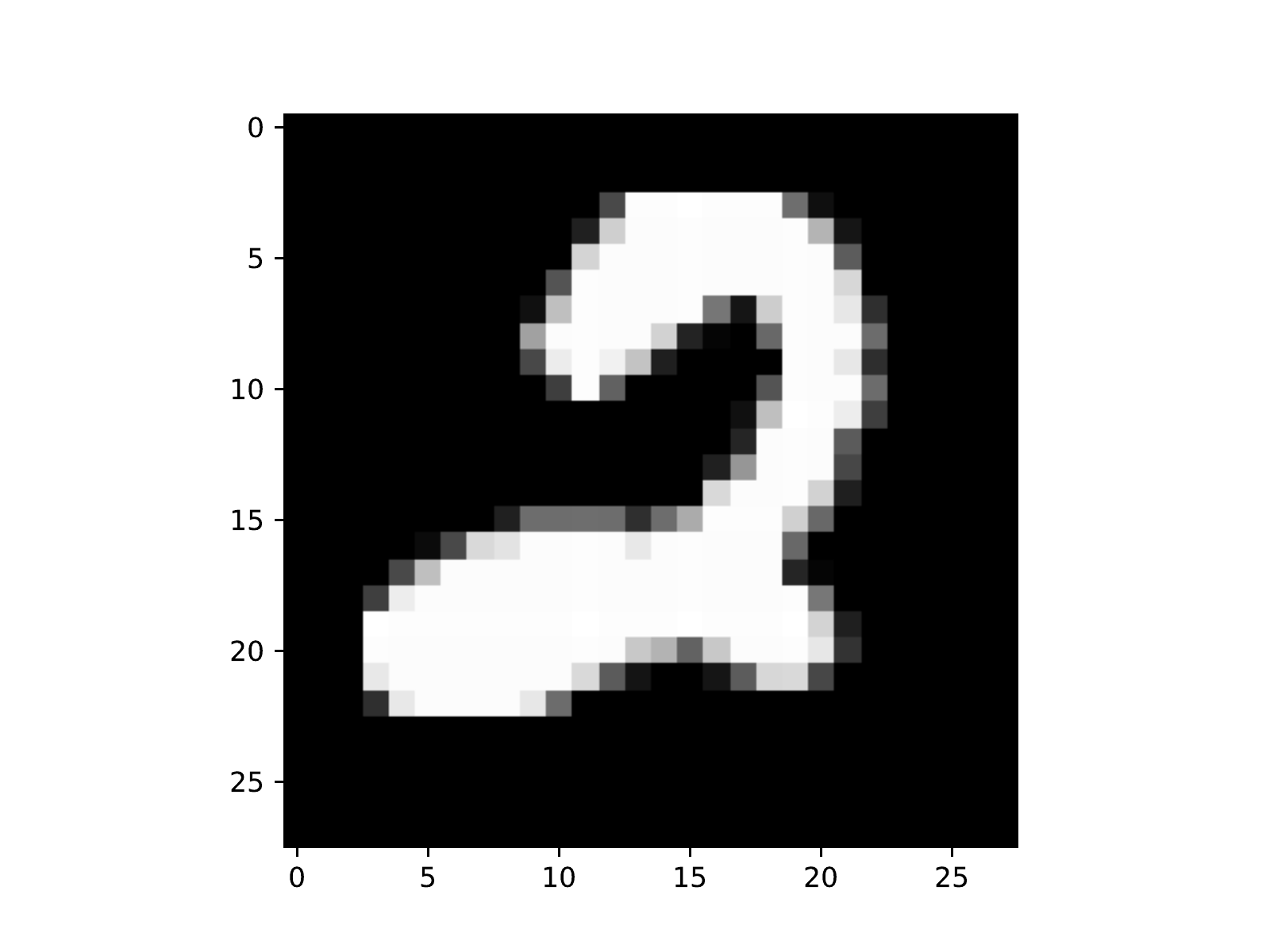}
\centerline{(b) "2" before Back-gradient}
\end{minipage}
\begin{minipage}[t]{0.23\textwidth}
\centering
\includegraphics[width=4.5cm]{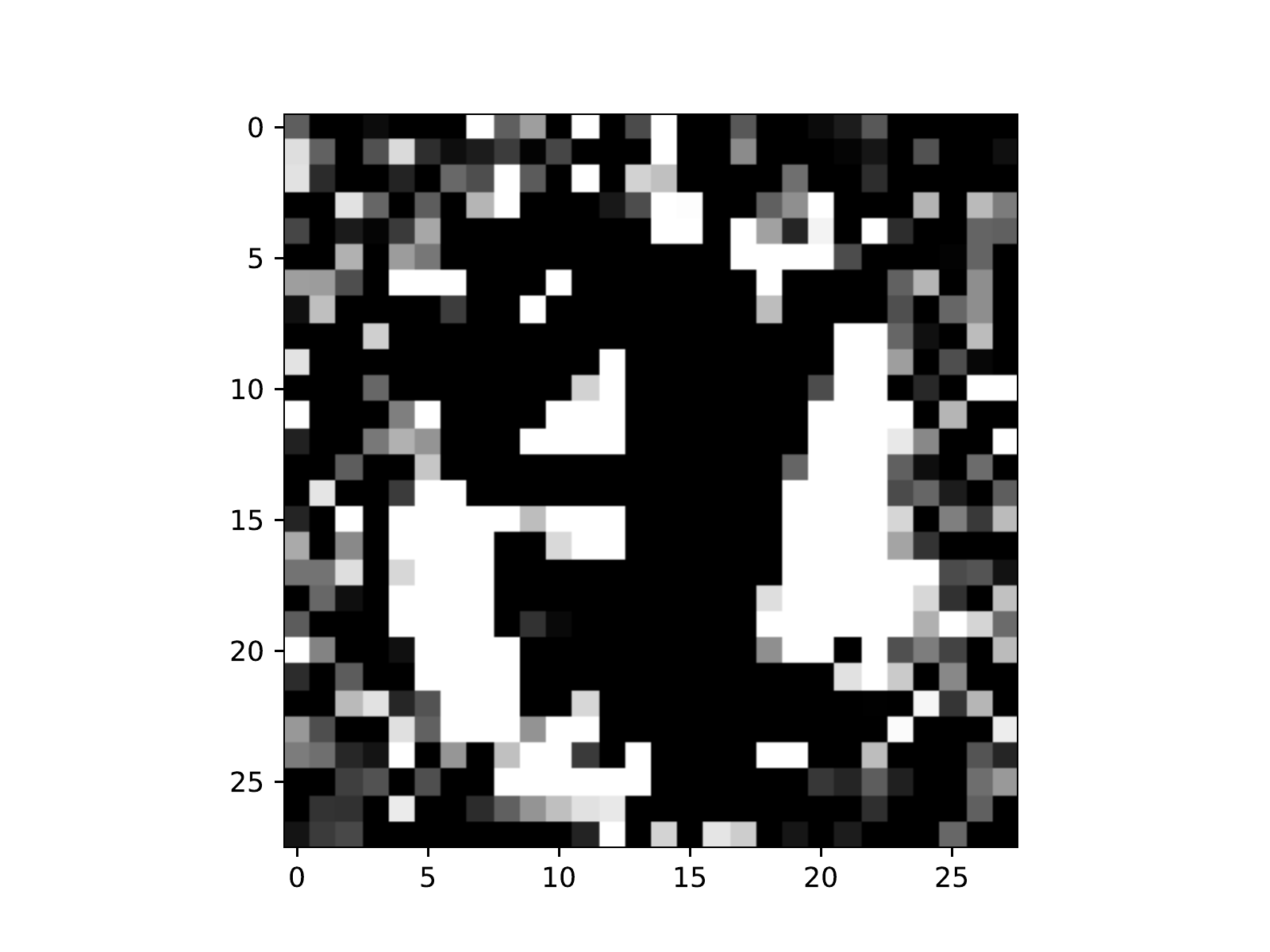}
\centerline{(c) "0" after Back-gradient}
\end{minipage}
\begin{minipage}[t]{0.23\textwidth}
\centering
\includegraphics[width=4.5cm]{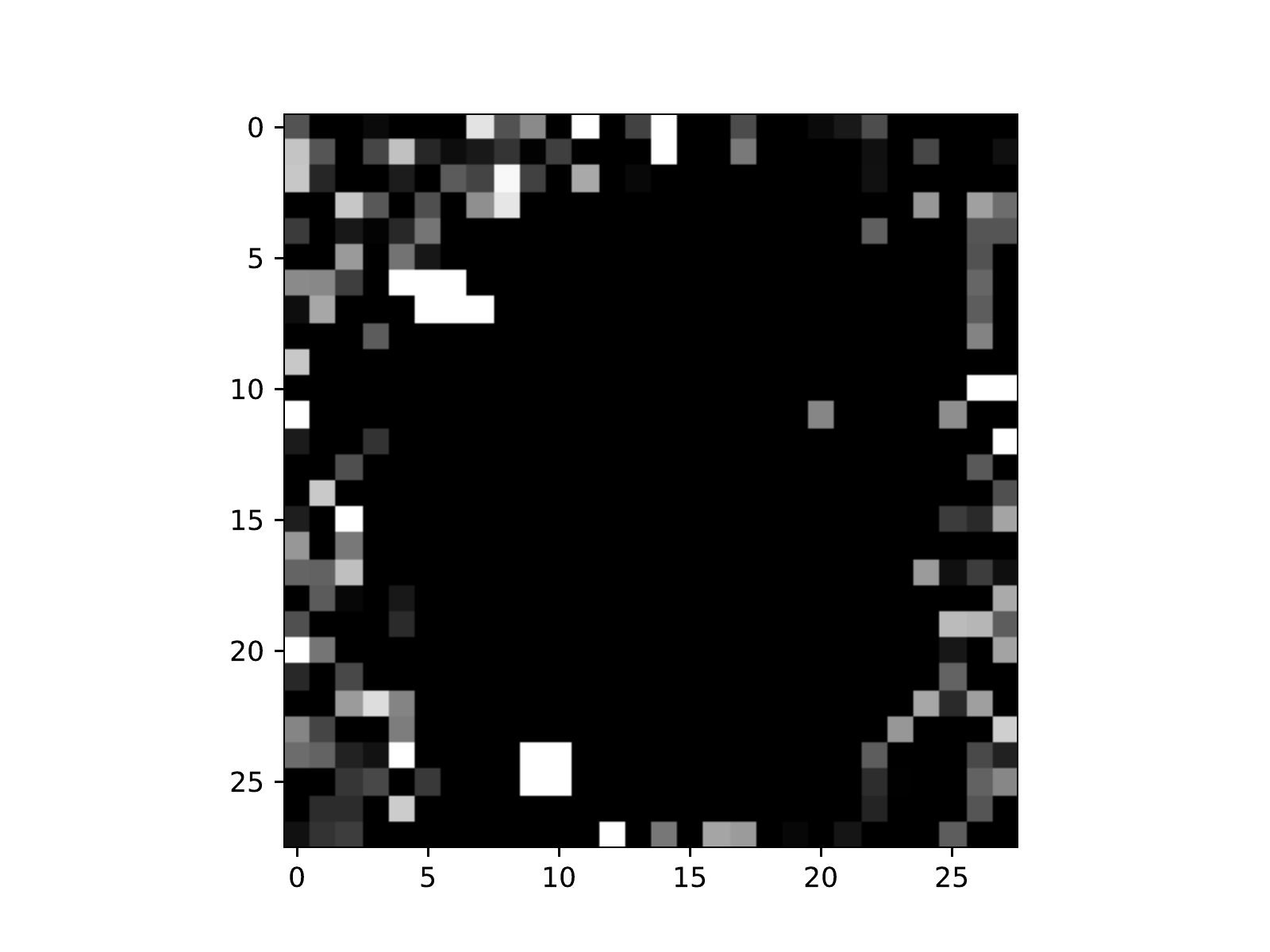}
\centerline{(d) "2" after Back-gradient}
\end{minipage}
\caption{Examples of poisoning samples made by Back-gradient attack algorithm.}
\label{sample_backgradient}
\end{figure}

\begin{figure*}[t]
\centering
\begin{minipage}[t]{0.24\textwidth}
\centering
\includegraphics[width=4.8cm]{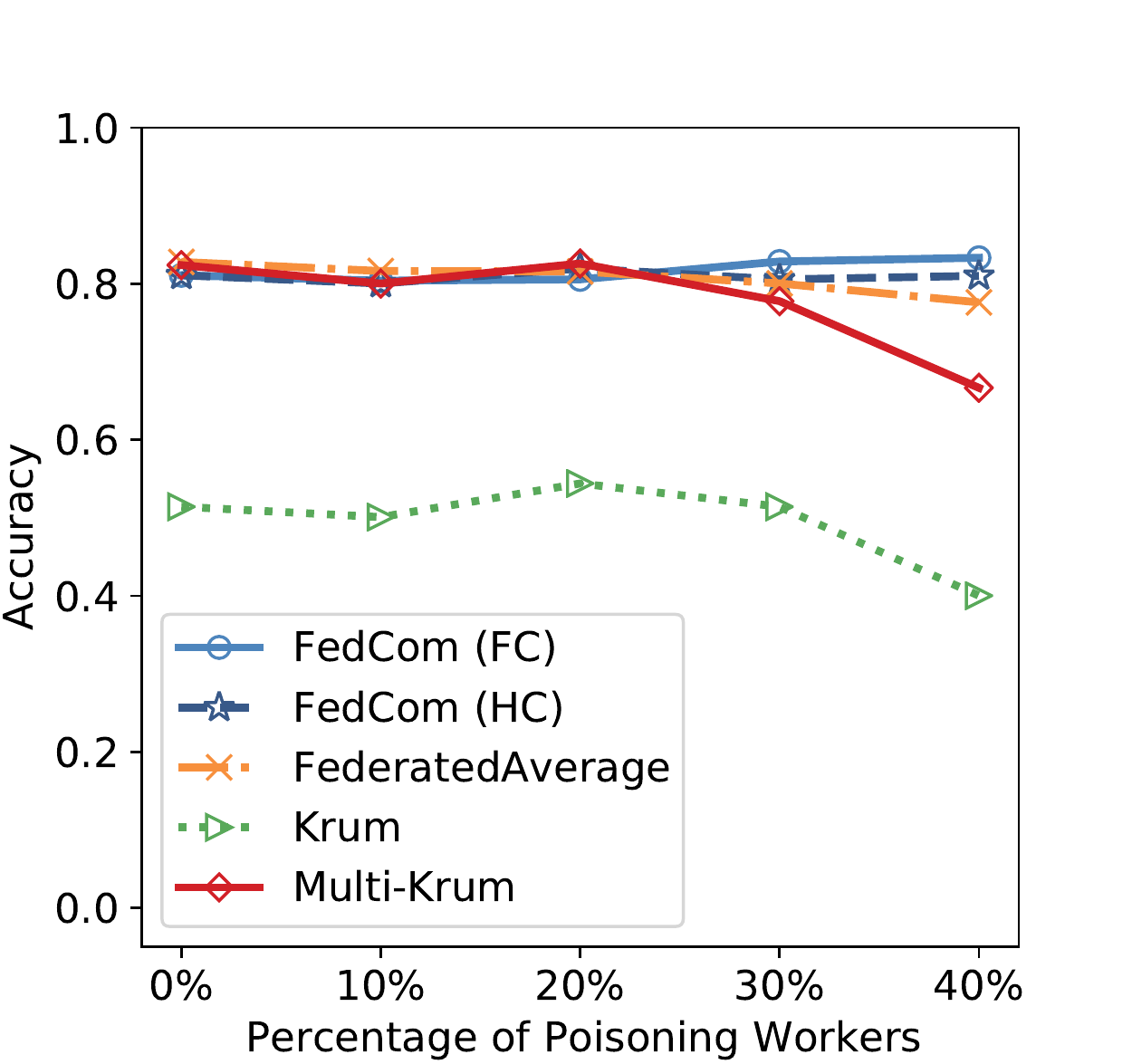}
\centerline{(a) MNIST}
\end{minipage}
\begin{minipage}[t]{0.24\textwidth}
\centering
\includegraphics[width=4.8cm]{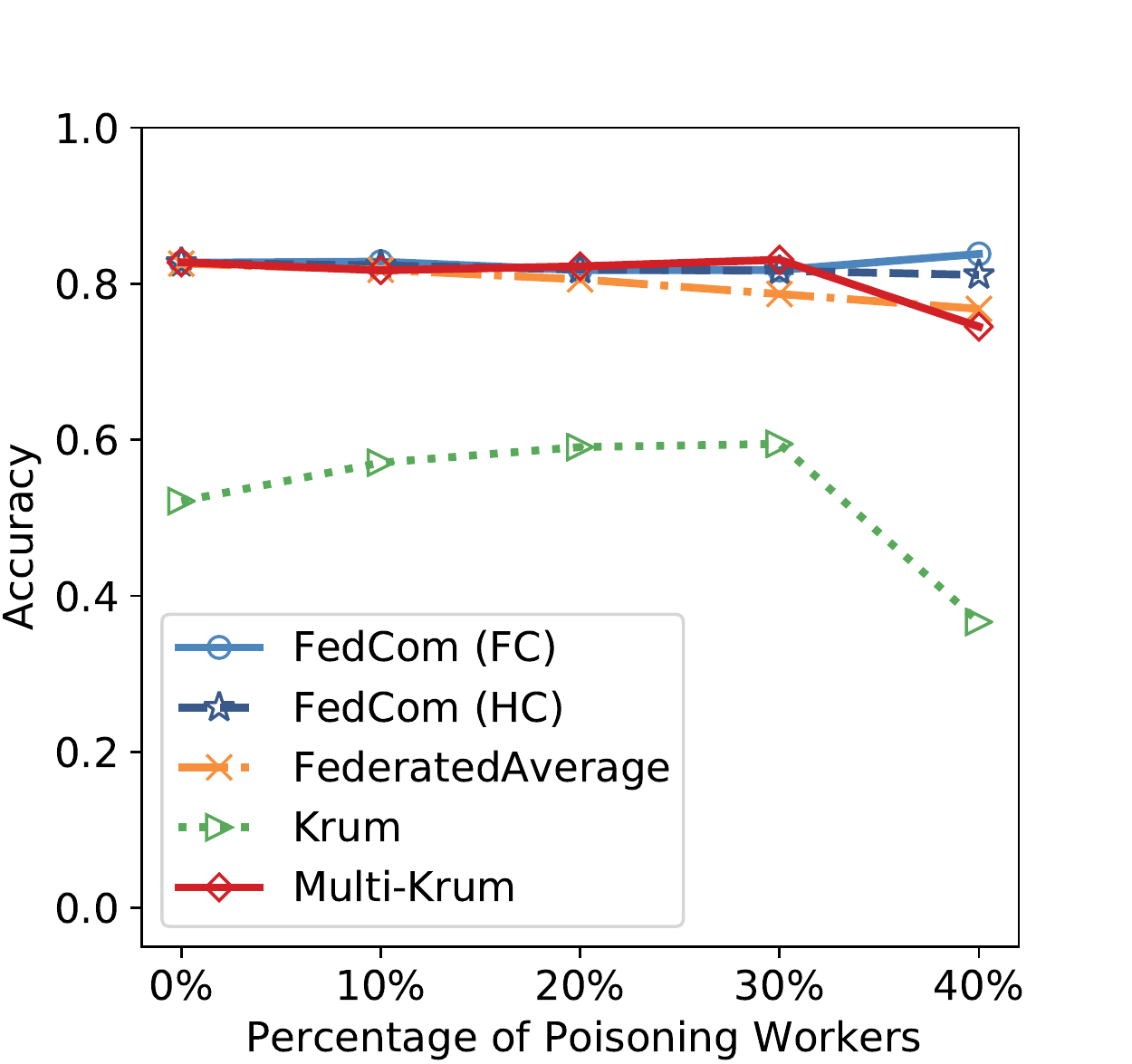}
\centerline{(b) MNIST}
\end{minipage}
\begin{minipage}[t]{0.24\textwidth}
\centering
\includegraphics[width=4.8cm]{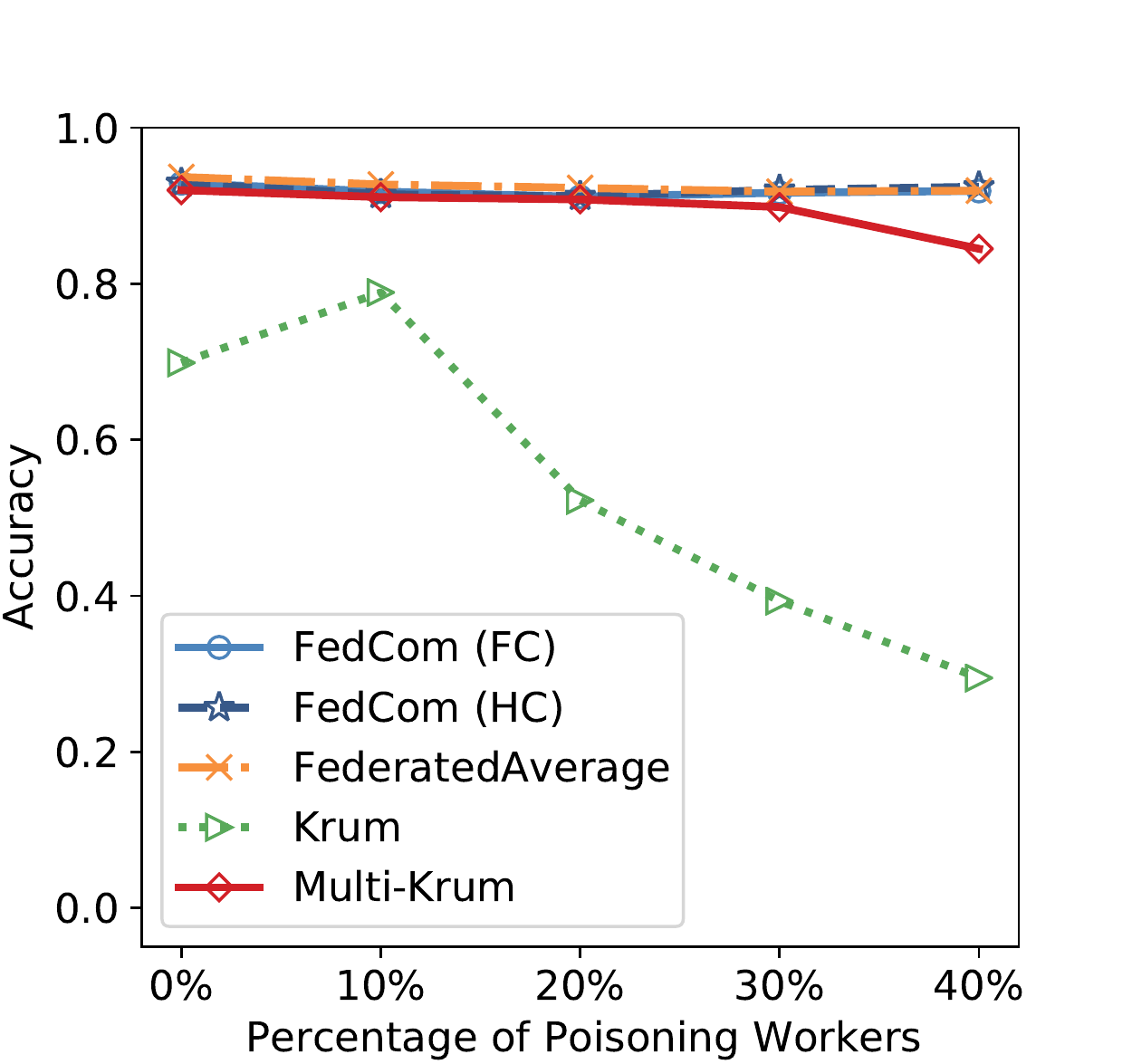}
\centerline{(c) HAR}
\end{minipage}
\begin{minipage}[t]{0.24\textwidth}
\centering
\includegraphics[width=4.8cm]{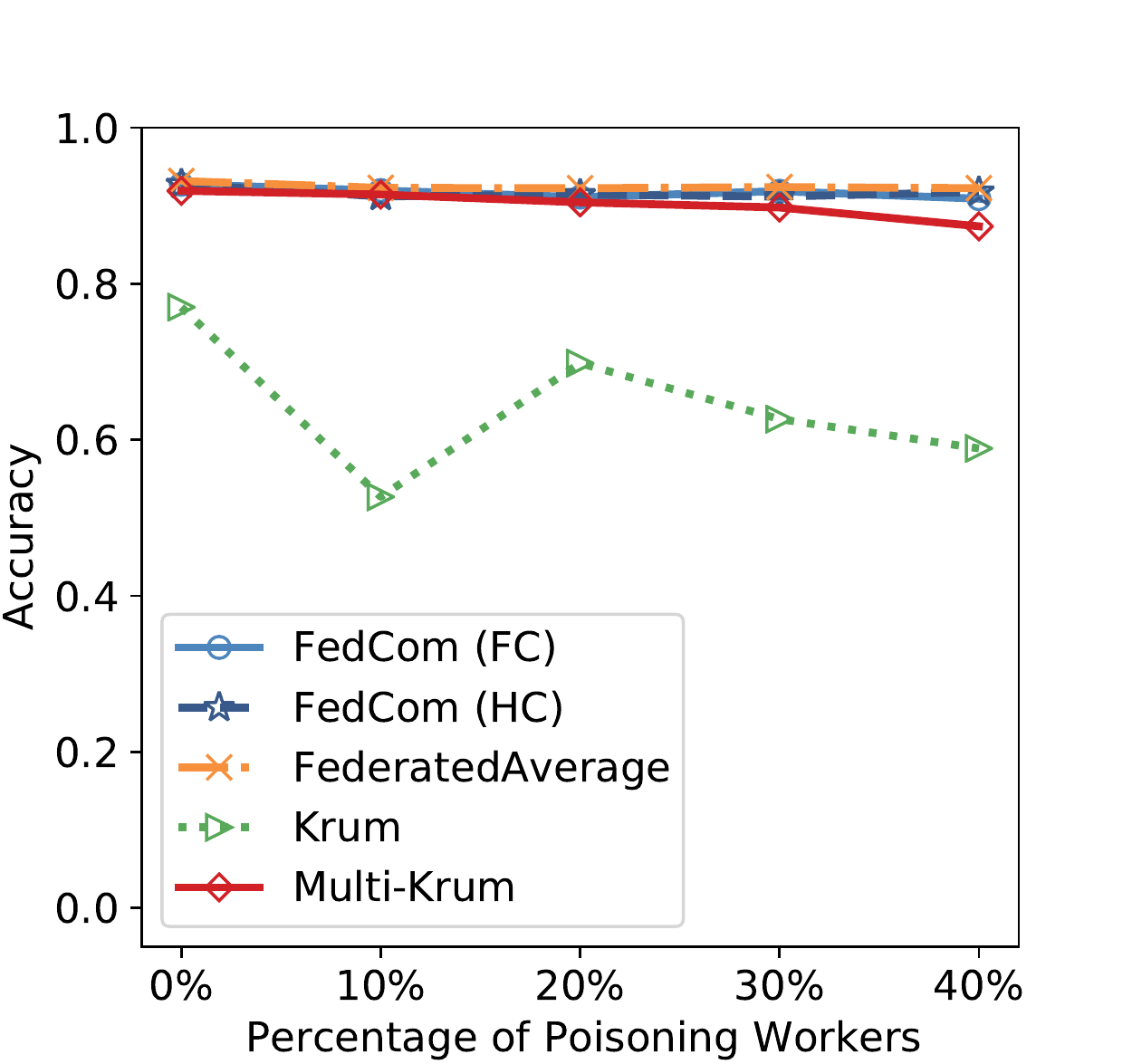}
\centerline{(d) HAR}
\end{minipage}
\begin{minipage}[t]{0.24\textwidth}
\centering
\includegraphics[width=4.8cm]{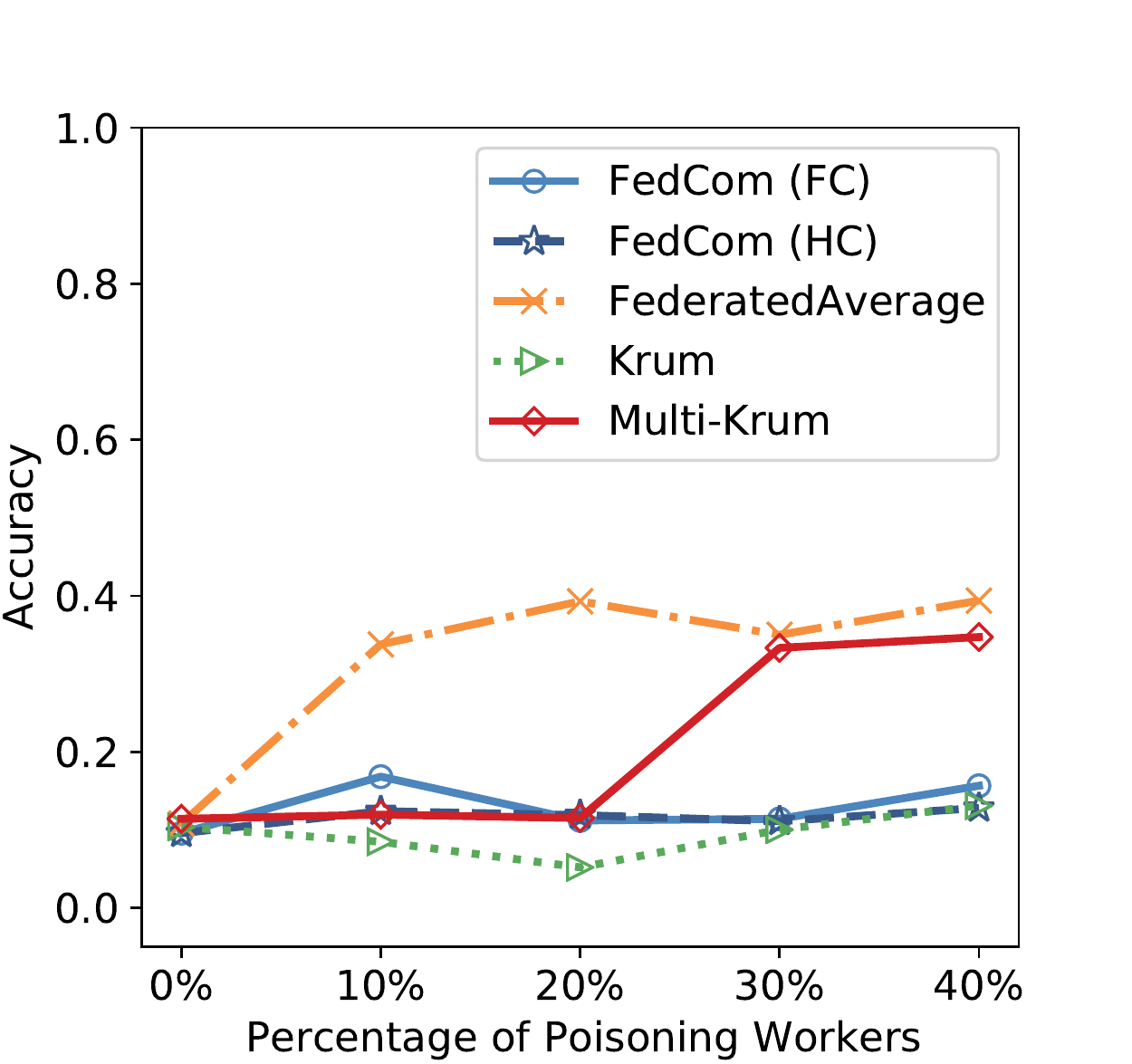}
\centerline{(e) MNIST}
\end{minipage}
\begin{minipage}[t]{0.24\textwidth}
\centering
\includegraphics[width=4.8cm]{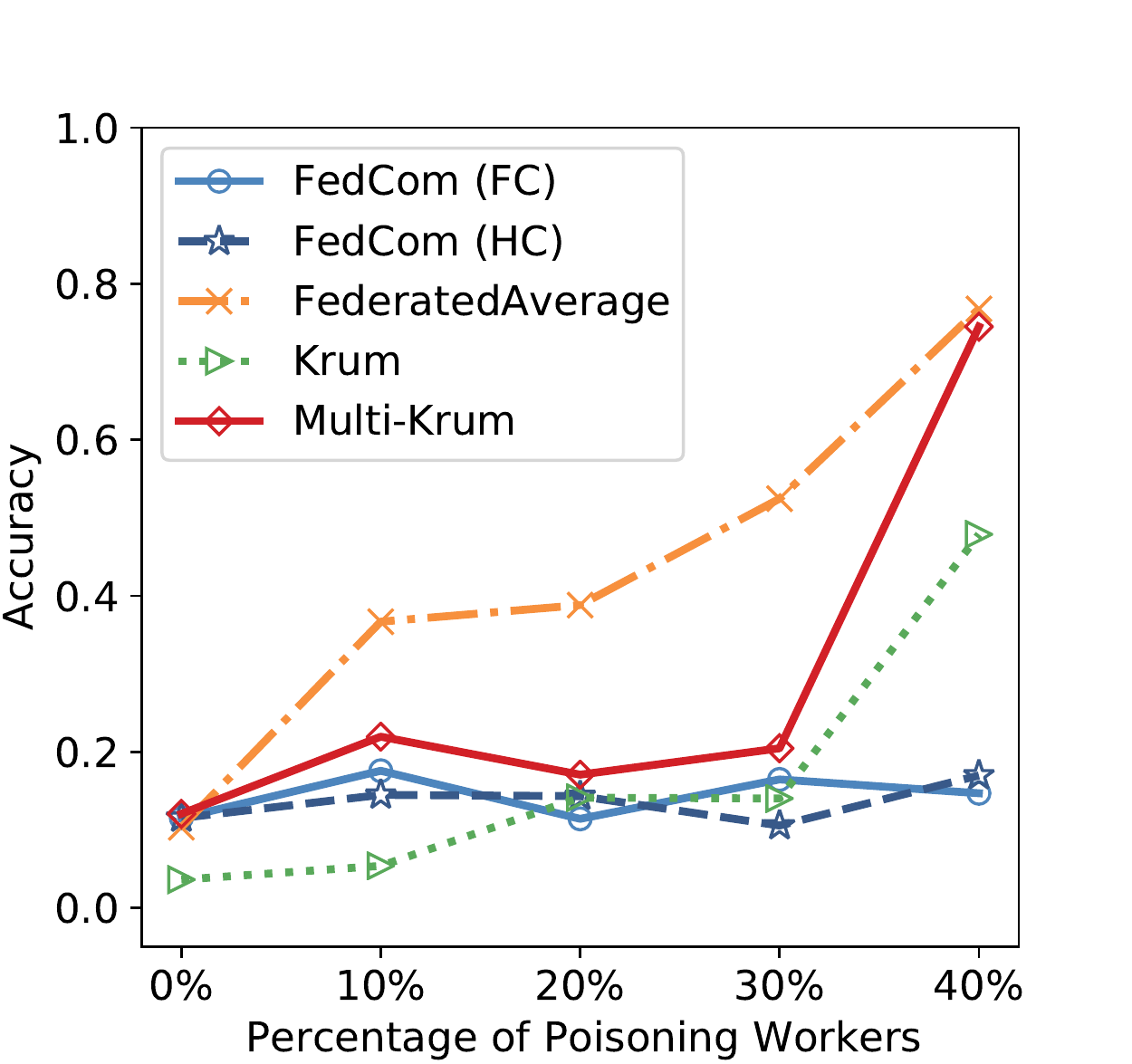}
\centerline{(f) MNIST}
\end{minipage}
\begin{minipage}[t]{0.24\textwidth}
\centering
\includegraphics[width=4.8cm]{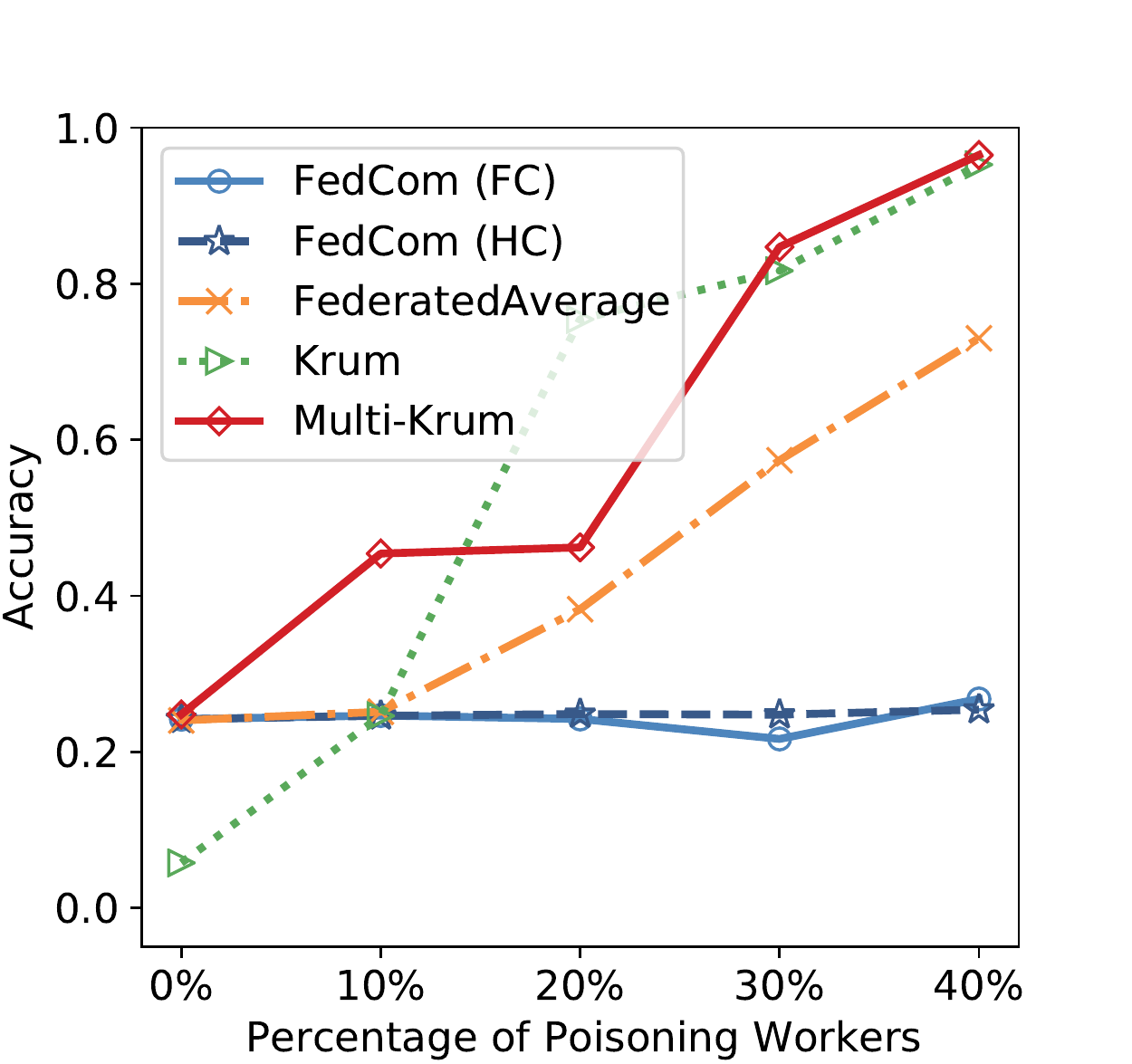}
\centerline{(g) HAR}
\end{minipage}
\begin{minipage}[t]{0.24\textwidth}
\centering
\includegraphics[width=4.8cm]{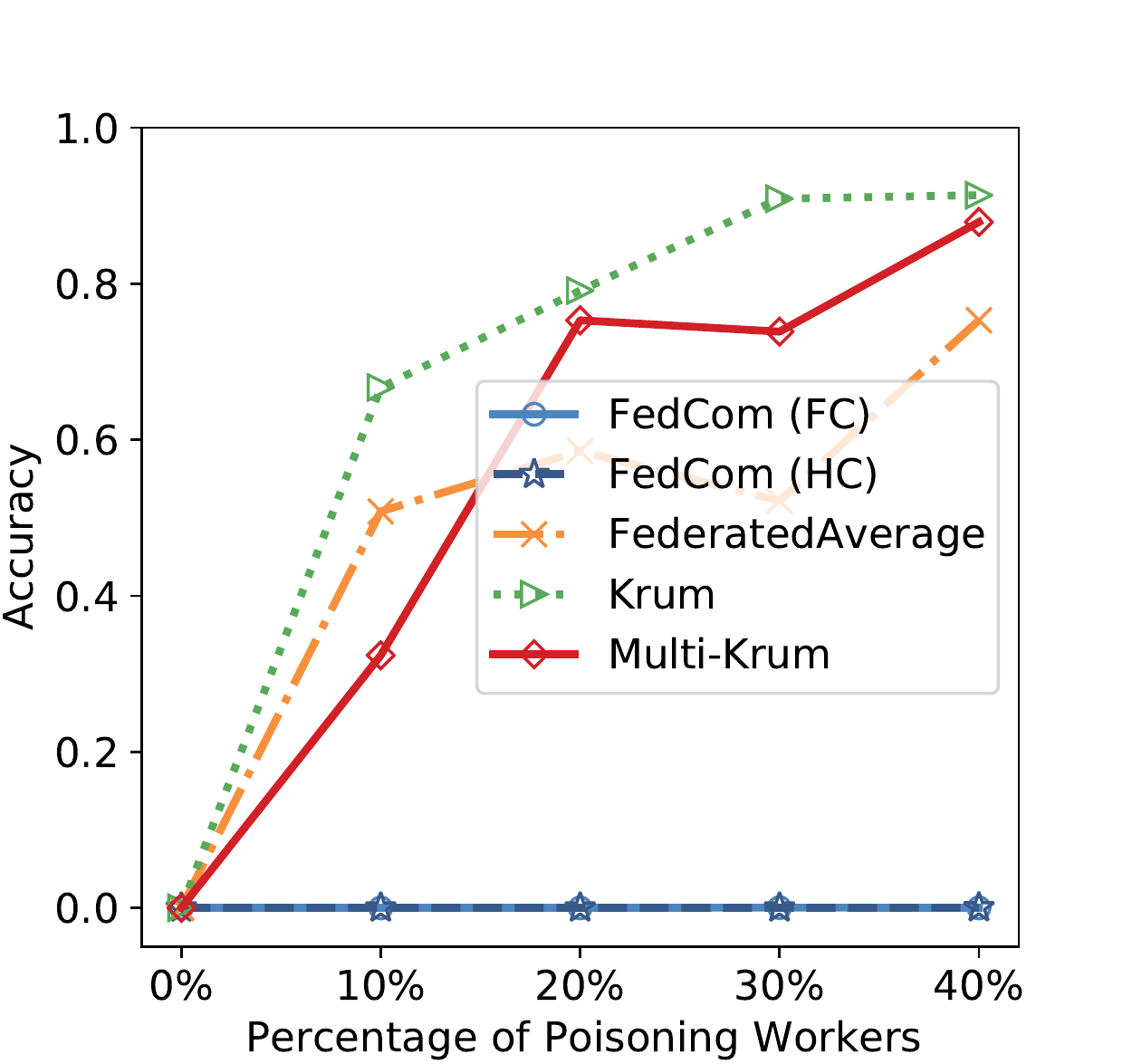}
\centerline{(h) HAR}
\end{minipage}
\caption{Results of model accuracy on benign and poisoning datasets under Back-gradient attack. (a) and (c) are results of LR on benign datasets, while (b) and (d) are result of DNN on benign datasets. (e) and (g) are results of LR on back-gradient poisoning datasets, while (f) and (h) are results of DNN on back-gradient poisoning datasets.}
\label{result_backgradient}
\end{figure*}

To further compare the performance of FedCom and other aggregation rules, we implemented advanced poisoning attacks, These attacks are more crafty compared with naive poisoning attacks above, and obtain more pertinence or stealthiness. We implemented Back-gradient attack as LDP, and Krum attack as LMP, and compete between these attacks and four aggregation rules. Figure \ref{result_krumattack} shows the results of Krum attack and Figure \ref{result_backgradient} shows the results of Back-gradient attack. We will discuss about such results from aspects below.

\textbf{Impact of Krum attack}: First of all, we could easily summarize that under \texttt{Krum} or \emph{Multi-}\texttt{Krum}, Krum attack could significantly reduce the accuracy of global models on benign validation set, hence, the Krum attack we implement is correct. Krum attack is specifically aimed at similar schemes like \emph{Multi-}\texttt{Krum} or \texttt{Krum}, but Krum attack is also claimed to obtain generality to \texttt{FedAverage}. Our evaluation also prove that Krum attack could also reduce the performance of global model under \texttt{FedAverage}. The results are also indicate that
a higher level of Non-IID will brings a better effectiveness of Krum attack, and such phenomenon meets the expectation of Krum attack’s original design. Concretely, Krum attack's effectiveness on HAR is relatively weaker, but we consider this is the result of lower degree of Non-IID that could make it easier for benign local models to correct the manipulated direction of global model. In a word, Krum attack's effectiveness meets our expectation. However, the results indicates that Krum attack is almost ineffective to FedCom, as global model's accuracy on benign validation set maintains the same high level as when there is no attack. Hence, we could consider that FedCom obtains better robustness than schemes like \emph{Multi-}\texttt{Krum} or \texttt{Krum}.

\textbf{Impact of Back-gradient attack}: in original Back-gradient attack, attacker need to randomly select samples from benign dataset as seeds, and flip their labels before processing these samples using back-gradient algorithm. In other word, the original Back-gradient attack is a Label-flipping-based attack, which adopts crafty back-gradient optimization algorithm to amplify the effectiveness of poisoning samples. We believe that such process is conspicuous, just like Label-flipping attack, as directions of poisoning local models trained by these poisoning datasets will be definitely opposite to that of correct global model. To weaken such conspicuousness, and distinguish Back-gradient attack from Label-flipping attack, we maintain the original labels of seeds in this attack. This is to ensure poisoning samples obtain furthest biases from original data distribution, and avoid directly "crossing" the decision boundary. We take MNIST as example and show partial poisoning samples in Figure \ref{sample_backgradient}. These examples intuitively indicates that the Back-gradient attack could significantly change the data distribution. The results in Figure \ref{result_backgradient} indicates that compared with Label-flipping attack, such an attack could successfully disable \emph{Multi-}\texttt{Krum} or \texttt{Krum}, as global models under these aggregation rules obtains decay of accuracy on benign validation set, as well as rise of accuracy on back-gradient poisoning datasets. \texttt{FedAverage} is with no exception. The effectiveness of attack could amplify with a rising percentage of poisoning workers, which also meets the expectation. We could also reach that Back-gradient attack is more destructive on MNIST. Concretely, the accuracy on poisoning dataset could indicate the degree global model fits poisoning data distribution, and it is obvious that model on HAR fits poisoning distribution better, but obtain smaller performance decay than model on MNIST. In other word, global model on MNIST may obtain worse performance even with a smaller degree of overfitting poisoning data distribution. Hence, Back-gradient attack could be more effective with higher degree of Non-IID. However, no matter whether attackers submit an honestly constructed commitment, and no matter what the degree of Non-IID or percentage of poisoning workers is, the accuracy of global model under FedCom maintains a similar high level as when there is no attack, and the rise of that on Back-gradient poisoning set could be neglected.

Through this part of evaluation, we could conclude that, \textbf{facing advanced poisoning attacks like Krum attack or Back-gradient attack, \emph{Multi-}\texttt{Krum} or \texttt{Krum} obtain no robustness, and there is a positive correlation between the effectiveness of attacks, and the degree of Non-IID or the percentage of poisoning workers, but FedCom obtains well robustness and tolerance to above attacks and the effect of Non-IID local datasets.}

\subsection{Summary}
Through our evaluation, we could conclude that FedCom obtains broader robustness than representative Byzantine robust aggregation rules like \emph{Multi-}\texttt{Krum} or \texttt{Krum}, as FedCom could be effective to both LDP and LMP, and maintain such effectiveness when attacks are advanced, while \emph{Multi-}\texttt{Krum} or \texttt{Krum} is disabled. Moreover, FedCom is also evaluated that its robustness will not decay with higher degree of Non-IID, which \emph{Multi-}\texttt{Krum} or \texttt{Krum} and their related schemes are proved to be more fragile in such case. Besides, FedCom obtains similar performance to \texttt{FedAverage}, \emph{Multi-}\texttt{Krum} or \texttt{Krum} when there is no attack, which means FedCom does not take the decay of ideal performance as price of such a broader robustness. We have summarize the contribution of FedCom into Table \ref{tabel_comparison}.

\begin{table}[]
    \caption{Comparison of Evaluated Aggregation Rules}
    \vspace{5pt}
    \centering
    \begin{tabular}{lcccc}
        \hline
         & FedCom & \texttt{Krum} & \emph{Multi-} & \texttt{Fed-} \\
         &        &               & \texttt{Krum} & \texttt{Average} \\
        \hline
        No attack  & $\surd$   & $\surd$ & $\surd$ & $\surd$ \\
        Gaussian  & $\surd$   & $\surd$ & $\surd$ & $\times$ \\
        Label-flip  & $\surd$   & $\surd$ & $\surd$ & $\times$ \\
        Krum  & $\surd$   & $\times$ & $\times$ & $\times$ \\
        Back-gradient  & $\surd$   & $\times$ & $\times$ & $\times$ \\
        Non-IID  & $\surd$   & $\times$ & $\times$ & $\times$ \\
        \hline
    \end{tabular}
    \label{tabel_comparison}
\end{table}

\section{Related Work}
\label{s5}
The privacy and secure problem of FL \cite{Advances_and_Open_Problems} attracts widespread attention recently, and many related works are proposed, many of which are closely related to our work.
\textbf{Poisoning attack}: in FL, poisoning attacks aim to obtain wrong global models via manipulating local training datasets (LDP) or local models (LMP), and try to make users use such models to perform inference tasks \cite{Canmachinelearningbesecure}. LDP has been much explored, which usually launch attack by constructing malicious samples into training dataset \cite{PoisoningAttacksagainstSupportVectorMachines, TargetedBackdoorAttacksonDeepLearning, PoisoningAttackstoGraph-BasedRecommenderSystems, ManipulatingMachineLearning:Poisoning, BadNets:IdentifyingVulnerabilitiesintheMachineLearningModel, PoisonFrogs!TargetedClean-LabelPoisoning}. LDP could be divide into targeted LDP and untargeted LDP, in which untargeted LDP \cite{PoisoningAttacksagainstSupportVectorMachines, DataPoisoningAttacksonFactorization-BasedCollaborative, ManipulatingMachineLearning:Poisoning} is to make global model perform worse on general validation set, while targeted LDP \cite{HowToBackdoorFederatedLearning, AnalyzingFederatedLearningthroughanAdversarialLens, ExploitingMachineLearningtoSubvertYourSpamFilter} is to make global model perform worse on specific subtasks, and perform normally on other subtasks. Some of these targeted LDP attacks are also known as \emph{backdoor attacks}. LMP is not wildly explored, but recently proposed work \cite{LocalModelPoisoningAttacksto} gives novel idea of jamming the training process of FL, even if the aggregation rule is claimed to be Byzantine-robust. Note that backdoor attacks in FL \cite{HowToBackdoorFederatedLearning} could also be sorted into LMP, as local models with backdoors needs to be trained by manipulated local datasets, but such an attack also requires a certain strategy of model replacement, which is to inject local model with backdoor into model aggregation when global loss is almost convergent.

\textbf{Distance Statistical Aggregation (DSA)}: we usually assume that more that half nodes in a distributed system is honest, as for an attacker, controlling or injecting more than half of malicious nodes is difficult. Some Byzantine-robust aggregation rules concretize such assumption: benign local model holds majority in all local models, and distance between two benign models is smaller than that between benign one and poisoned one, because benign local models are honestly trained on similar local datasets. \texttt{Krum} \cite{Machine_Learning_with_Adversaries} is one of representative aggregation rules based on such an assumption. In \texttt{Krum}, each local model $w_{i}$ obtains a score based on distances between $w_{i}$ and other local models. Concretely, \texttt{Krum} firstly calculates the sum of Euclidean distances between $w_{i}$ and the $n-f-1$ closest local models, $n$ is the total number of workers while $f$ is the upper bound of Byzantine workers number. \texttt{Krum} works if $f<(n-2)/2$. \texttt{Krum} will select the local model with lowest score as global model in one iteration. \texttt{Krum} always select local model which is the closest to the centroid of majority model cluster. In high dimension spaces, crafting poisoning local model by changing few parameters in local model could constraint a small distance between poisoning model and the correct one, which \texttt{Krum} may not be able to handle with. \texttt{Trimmed Mean} \cite{Byzantine-RobustDistributedLearning:Towards} is proposed to against such attacks. Such an aggregation rule checks the value of each parameter in global model. \texttt{Trimmed Mean} sorts the $i$th parameters of all local models, and remove $f$ biggest and smallest parameters, take the mean of left parameters as the $i$th parameter of global model. \texttt{Median} \cite{Byzantine-RobustDistributedLearning:Towards} is similar to \texttt{Trimmed Mean}, but takes medians as parameters of the global model.

Local datasets in federated learning are usually Non-IID, which breaks the assumption of schemes mentioned above, and introduce vulnerabilities. Fang et al \cite{LocalModelPoisoningAttacksto} exploited such vulnerabilities, and successfully impact the robustness of \texttt{Krum}, \texttt{Bulyan}, \texttt{Trimmed Mean} and \texttt{Median}. The basic idea of their attacks is attempting to craft a poisoning local model, and this model obtains both minimum distances to other models, and deviated direction from correct global gradient. Such wrong directed crafted model could always be chosen by Krum or other schemes as global model.

\textbf{Contribution Statistical Aggregation (CSA)}: there is also an assumption that honest workers could always submit local models that positively contribute to global model's convergence. Concretely, a benign local model could positively impact global model accuracy, or negatively impact global model loss on certain validation set, if taken as new global model. Such assumptions derive two types of aggregation rules: loss function based rejection (LFR) and error rate based rejection (ERR) \cite{LocalModelPoisoningAttacksto}. Representative schemes adopting ideas above are \texttt{RONI} \cite{Thesecurityofmachinelearning}, \texttt{Zeno} \cite{Zeno:DistributedStochasticGradientDescent} and \texttt{Zeno++} \cite{Zeno++:RobustFullyAsynchronous}. \texttt{RONI} determine each sample's contribution to accuracy of global model, and reject samples that introduce negative contributions. Such an idea could be generalized to eliminating poisoning local models. \texttt{Zeno} determine loss contributions of local models instead.

Obviously, in distributed learning that IID sampling is not restricted, such schemes could isolate poisoning attack from the underlying level. However, such schemes could be infeasible in federated learning, because they depends on an effective public validation set to determine contribution of local models. First of all, forming such a dataset by IID sampling violates the principle of federated learning that user's training dataset remains well-preserved privacy. Secondly, forming such a dataset by workers' negotiation about public knowledge could be difficult, as workers could only reach consensus on the "overlapping" parts of local datasets, but Non-IID data circumstance in federated learning makes local datasets "overlap" slightly, or even completely "separated". In conclusion, the restriction of sampling and Non-IID data circumstance of federated learning could make schemes based on ERR of LFR ineffective.

\textbf{Other defenses of poisoning attack}: in other machine learning system that third-parties have access to training data, poisoning attacks may be defeated by checking training dataset \cite{RobustLogisticRegressionandClassification, ManipulatingMachineLearning:Poisoning, RobustLinearRegressionAgainstTrainingDataPoisoning, CertifiedDefensesforDataPoisoningAttacks}. Obviously, such methods are not applicable in FL, as local datasets are privacies of workers.

\section{Conclusion}
\label{s6}

In this paper, we borrowed the idea of commitment from cryptography, and proposed a data commitment based Byzantine-robust FL framework which could simultaneously defend against data poisoning and model poisoning attacks even when data among workers is non-IID. By requiring each involved worker to submit a data commitment via sample cluster averaging to her private training data, the proposed FedCom could defend against data poisoning attacks by comparing the Wasserstein distance among the data commitments, and thwart model poisoning attacks by evaluating the behavior of different local models using the corresponding data commitment and conducting a behavior-aware local model aggregation. We conduct an exhaustive performance evaluation of FedCom. The experimental results demonstrate that FedCom outperforms the state-of-the-art Byzantine-robust FL schemes in defending against typical data poisoning and model poisoning attacks under practical non-IID data distribution among workers. Note that we mainly focus on non-targeted attacks in this work, while in the future work we will investigate how to achieve robust FL when backdoor (targeted) poisoning attacks happen.

\section*{Acknowledgment}
We thank the anonymous reviewers for precious comments, and Yongdeng Zhao for his kindly donation and comments.



%
%
%

\bibliography{conference_101719}
\bibliographystyle{IEEEtran}

\end{document}